\documentclass{article} 
\usepackage{conference,times}

\usepackage{amsmath,amsfonts,bm}

\def\eqref#1{equation~\ref{#1}}

\def\1{\bm{1}}

\DeclareMathAlphabet{\mathsfit}{\encodingdefault}{\sfdefault}{m}{sl}
\SetMathAlphabet{\mathsfit}{bold}{\encodingdefault}{\sfdefault}{bx}{n}

\usepackage{hyperref}
\usepackage{url}
\usepackage{graphicx} 

\definecolor{lightgray}{RGB}{220, 220, 220}
\definecolor{basedarkblue}{RGB}{0,0,80}     
\colorlet{darkblue}{basedarkblue!75}
\colorlet{lightblue}{basedarkblue!8}

\definecolor{burntorange}{RGB}{204,85,0}  
\colorlet{lightburntorange}{burntorange!8}

\definecolor{basecream}{RGB}{255,240,190}
\colorlet{cream}{basecream!70!black}
\colorlet{lightcream}{cream!8}

\definecolor{basemaroon}{RGB}{160,0,0}
\colorlet{maroon}{basemaroon!70!black}
\colorlet{lightmaroon}{maroon!8}

\usepackage[most]{tcolorbox}
\usepackage{array}
\usepackage{xcolor}
\usepackage{wrapfig}
\usepackage{multirow}
\usepackage{subcaption}
\usepackage{booktabs}
\definecolor{headerblue}{RGB}{70, 90, 120}
\definecolor{lightgray}{RGB}{220, 220, 220}

\title{Opponent Shaping in LLM Agents}

\author{Marta Emili Garcia Segura\textsuperscript{1, 2}, Stephen Hailes\textsuperscript{1}, Mirco Musolesi\textsuperscript{1, 2, 3} \\ 
\textsuperscript{1}Department of Computer Science, University College London\\
\textsuperscript{2}Centre for Artificial Intelligence, University College London  \\
\textsuperscript{3}Department of Computer Science, University of Bologna \\
\texttt{\{marta.segura22, s.hailes, m.musolesi\}@ucl.ac.uk}
}

\iclrfinalcopy 
\begin{document}

\maketitle

\begin{abstract}
Large Language Models (LLMs) are increasingly being deployed as autonomous agents in real-world environments. As these deployments scale, multi-agent interactions become inevitable, making it essential to understand strategic behavior in such systems. A central open question is whether LLM agents, like reinforcement learning agents, can shape the learning dynamics and influence the behavior of others through interaction alone. In this paper, we present the first investigation of opponent shaping (OS) with LLM-based agents. Existing OS algorithms cannot be directly applied to LLMs, as they require higher-order derivatives, face scalability constraints, or depend on architectural components that are absent in transformers. To address this gap, we introduce ShapeLLM, an adaptation of model-free OS methods tailored for transformer-based agents. Using ShapeLLM, we examine whether LLM agents can influence co-players’ learning dynamics across diverse game-theoretic environments. We demonstrate that LLM agents can successfully guide opponents toward exploitable equilibria in competitive games (Iterated Prisoner’s Dilemma, Matching Pennies, and Chicken) and promote coordination and improve collective welfare in cooperative games (Iterated Stag Hunt and a cooperative version of the Prisoner’s Dilemma). Our findings show that LLM agents can both shape and be shaped through interaction, establishing opponent shaping as a key dimension of multi-agent LLM research.
\end{abstract}

\section{Introduction}

Large language models (LLMs) have evolved rapidly in recent years, demonstrating remarkable capabilities in reasoning, planning and goal-directed behavior that make them increasingly suitable for deployment as autonomous agents \citep{zhao2023survey, anthropic2025opus4, openai2025gpt5, Xi2025therise}. Already, LLM-based agents are being adopted for complex tasks such as web navigation and code generation \citep{anthropicClaudeCode, openAIoperator}. As deployment scales, these agents will be less likely to operate in isolation. Instead, they will increasingly interact with other learning agents in shared environments, collaborating on tasks, competing for resources, or pursuing independent objectives. There is growing interest in understanding the opportunities and challenges associated with multi-agent LLM systems \citep{fourney2024magentic, ghafarollahi2025sciagents, pan2025AgentCoord, rosser2025AgentBreeder}. However, most approaches treat LLMs as static entities, overlooking the strategic dynamics that emerge when agents continuously adapt to one another.

Multi-agent reinforcement learning (MARL) has long been concerned with the interaction of multiple learners in shared environments \citep{busoniu2008comprehensive}. A core difficulty in MARL is that agents often treat each other as static parts of the environment, which can yield poor collective outcomes. For instance, in the Iterated Prisoner’s Dilemma (IPD, \citet{axelrod1981evolution}), independent learners reliably converge to mutual defection, which is the worst collective outcome \citep{harper2017reinforcement, foerster2018LOLA}. To mitigate such failures, the field of opponent shaping develops agents that actively anticipate and influence their co-players’ learning dynamics, steering learned behavior toward more favorable equilibria \citep{foerster2018LOLA, lu2022mfos}. While these methods have proven effective in RL settings, it remains unclear whether they extend to LLM agents. Unlike RL agents, these models process rich semantic information, exhibit complex reasoning capabilities \citep{xu2025towards}, and can adapt through in-context learning \citep{brown2020language, bubeck2023sparks}. This raises uncertainty about whether traditional shaping methods will transfer to LLM agents. Moreover, existing algorithms are poorly suited to LLMs, as they rely on estimating higher-order derivatives, adopt a dual-agentic structure, or require architectural components unavailable in transformers \citep{foerster2018LOLA, lu2022mfos, khan2024shaper}.

Our work addresses this gap, being the first exploration of opponent shaping with LLM agents. Understanding the extent to which LLMs can engage in opponent shaping is critical as they are increasingly deployed in real-world, multi-agent settings. This capability has dual implications: LLM agents may be vulnerable to exploitation by adversaries who strategically influence their learning dynamics, while shaping could also be a tool to foster prosocial behavior and enable coordination.

We study whether LLM agents can strategically influence each other’s learning dynamics in repeated matrix games, which capture core strategic incentives while allowing precise outcome quantification. As a baseline, we train two LLM agents independently using Proximal Policy Optimization (PPO, \cite{schulman2017proximal}) to maximize their individual returns. We then introduce opponent shaping by turning one of those agents into a  \textit{shaper} that aims to alter the learning dynamics of its co-player. The shaper is trained using \textit{ShapeLLM}, our proposed algorithm that adapts model-free approaches \citep{lu2022mfos, khan2024shaper} to transformer architectures. We evaluate the efficacy of shaping in both exploitative scenarios, where an agent seeks unilateral advantage, and prosocial scenarios, where shaping fosters cooperation. Our contributions are the following:

\begin{itemize}

\item We provide the first investigation of opponent shaping in LLM agents, demonstrating that they can strategically influence each other's learning dynamics through interaction alone.

\item We propose \textit{ShapeLLM}, a model-free opponent shaping algorithm for transformer architectures leveraging structured natural language prompts.

\item We evaluate shaping across diverse game-theoretic environments and show that LLM agents can successfully exploit opponents in competitive settings and guide interactions toward mutually beneficial outcomes in cooperative ones.

\end{itemize}

\section{Background}
\label{background}

\subsection{LLM Agents}

An \textit{LLM agent} can be thought of as any system leveraging LLMs as the core computational unit for reasoning, planning, and decision-making \citep{sumers2023cognitive}. The architectures of these agents vary significantly in the literature, with some systems integrating reasoning frameworks \citep{yao2023react}, memory banks \citep{vezhnevets2023concordia}, or tool-access via APIs \citep{schick2023toolformer, patil2024gorilla}, both in single and multi-agent settings \citep{park2023generative, wang2024survey}. In the latter, game-theoretic environments provide a natural testbed for studying strategic dynamics. Recent work has used these environments to investigate LLM agents’ cooperation \citep{piatti2024cooperate,akata2025playing}, rationality \citep{fan2024can}, and strategic reasoning \citep{gandhi2023strategic, duan2024gtbench, huang2025far}. Beyond observational studies, these environments can be used to train LLMs towards specific objectives such as moral alignment \citep{tennant2025moral}.

\subsection{Fine-tuning LLMs with Reinforcement Learning}

The use of reinforcement learning (RL) in LLM training was first popularized through Reinforcement Learning from Human Feedback (RLHF, \citet{zeigler2019RLHF, stiennon2020learning, ouyang2022training}), which typically employs Proximal Policy Optimization (PPO, \citet{schulman2017proximal}) as the RL algorithm. PPO is an on-policy, actor-critic method that uses a learned value function for advantage estimation. When applied to LLMs, it is customary to include a Kullback-Leibler (KL) penalty in the reward signal to prevent the model's output distribution from diverging too far from the pre-trained one. Several alternatives to standard RLHF with PPO have since been proposed, including Group Relative Policy Optimization (GRPO, \citet{shao2024deepseekmath}), which estimates advantages via Monte Carlo rollouts, and Direct Preference Optimization (DPO, \citet{rafailov2023direct}), which converts the RLHF objective into a supervised learning loss. All three approaches are typically used in contextual bandit settings \citep{grattafiori2024llama, yang2025qwen3, rastogi2025magistral}, where each model response is treated as an independent episode with immediate reward feedback. The application of multi-turn RL to LLMs remains an active area of research due to challenges in preference collection, reward modeling, and ambiguity in action space definition \citep{shani2024multi, zhou2024archer, zeng2025reinforcing}. Nevertheless, underlying algorithms such as PPO are inherently designed to handle temporally structured environments, making them suitable for multi-agent strategic settings where actions have long-term consequences.

\subsection{Opponent Shaping}

The opponent shaping literature is characterized by two primary approaches: methods that explicitly account for the opponent's updates in the agent's learning rule \citep{foerster2018LOLA, letcher2019stable, willi2022cola}, and meta-learning approaches that learn to shape opponents by observing how their actions influence their opponent's parameter updates \citep{lu2022mfos, balaguer2022shepherd, khan2024shaper}. The most notable method in the first category is Learning with Opponent-Learning Awareness (LOLA, \citet{foerster2018LOLA}). In LOLA, the opponent's learning rule is incorporated into the agent's update, accounting for the effect of the agent's action on the opponent's parameter updates. This method has demonstrated notable successes, such as the emergence of Tit-for-Tat (TFT) in the iterated prisoner’s dilemma (IPD) through self-play. However, LOLA faces significant limitations: it assumes knowledge of the opponent's learning rule,  relies on high-variance higher order derivatives, and only considers immediate effects on the opponent's updates. While several LOLA refinements have been proposed \citep{letcher2019stable, willi2022cola}, they exhibit the same core limitations.

The second category of methods, exemplified by Model-Free Opponent Shaping (M-FOS, \citet{lu2022mfos}), was developed specifically to overcome these challenges. M-FOS bypasses these limitations, particularly LOLA's myopic perspective, by framing opponent shaping as a meta-learning problem. In doing so, it decouples the task of interacting with the environment from that of influencing the opponent's learning dynamics. This decoupling is achieved via a bi-level agent architecture: an inner agent that interacts with its co-players, and an outer agent that updates or conditions the inner agent's policy. The outer agent operates in a meta-game, where the meta-state consists of all players' parameters, and the meta-action determines the inner agent's policy. Between episodes, other players update their parameters using their respective learning algorithms. This formulation enables the meta-agent to optimize for long-term opponent shaping effects.

While M-FOS has demonstrated strong empirical results, including outperforming LOLA-based agents in the IPD, it presents scalability challenges due to its dual-agent architecture. To address these, \citet{khan2024shaper} propose SHAPER, which simplifies M-FOS' architecture by collapsing the shaping agent into a single recurrent neural network (RNN). The key insight is the distinction between history and context within opponent shaping. History captures intra-episode information necessary for implementing conditional strategies such as TFT, while context captures inter-episode information about the opponent's learning dynamics. SHAPER captures history through the RNN's inputs and context through its hidden state, which persists across episodes within a trial. This unified architecture eliminates the dual action spaces of M-FOS, allowing the agent to operate directly in the environment's original action space. However, SHAPER is inherently tied to RNN architectures, with its mechanism relying on distinct memory streams for capturing history and context.

\section{Methodology}
\label{methodology}

\subsection{Preliminaries}
Agents interact with the environment by generating text. Let $\mathcal{V}$ denote the model's vocabulary, and $w_{1:L} := (w_1, \ldots, w_L)$, with $w_l \in \mathcal{V}, \forall l \in \{1, \ldots, L\}$. At each interaction, the agent's action is a sequence of tokens sampled from: 
\begin{equation}
    \label{agent_policy}
    \rho_{\theta}(w_{1:L} \mid c) = \prod_{l=1}^{L} \rho_{\theta}(w_l \mid c, w_{<l}),
\end{equation}
where $ w_{<l} := w_{1:l-1} $, the context $c$ is the environment's description, and $L$ is the generation length. For simplicity, we set $L=1$ and define the distribution $\rho_\theta (w \mid c)$, with $w \in \mathcal{V}$, as our agent's policy. We deliberately avoid constrained decoding \citep{beurerkellner2024, ugare2024syncode} or format-specific fine-tuning, relying instead on textual instructions to guide the model toward the desired format.

We formalize our environments as \textit{repeated normal-form games} \citep{fudenberg1991game}. Let $\mathcal{M} = (I, \{\mathcal{A}_i\}_{i \in I}, \{R_i\}_{i \in I})$ denote a base game, where $I = \{1, \ldots n\}$ represents the set of players. Each player $i \in I$ has an associated action space $\mathcal{A}_i$ and reward function $R_i: \mathcal{A}_1 \times \ldots \times \mathcal{A}_n \to \mathbb{R}$. A repeated normal-form game consists of playing $\mathcal{M}$ for $T$ time steps, where $T$ can be finite or infinite. In this paper, we focus on the finite case. At each time step $t$, players simultaneously choose actions $a_i^t$ for $i \in I$. The resulting joint action $\mathbf{a}^t = (a_1^t, \ldots, a_n^t)$ determines the reward $r_i^t = R_i(\mathbf{a}^t)$ that each player receives. The actions $a_i^t$ are sampled from player-specific policies $\rho_{\theta_i}(w \mid f(h^t))$, where $f(h^t)$ is any function of the joint action history $h^t = (\mathbf{a}^1, \mathbf{a}^2, \ldots, \mathbf{a}^{t-1})$ (e.g., the previous joint action $\mathbf{a}^{t-1}$).

\subsection{Opponent Shaping in LLM Agents}
\label{sec:opp_shaping}

\begin{figure}[h]
\begin{center}
\includegraphics[width=\linewidth]{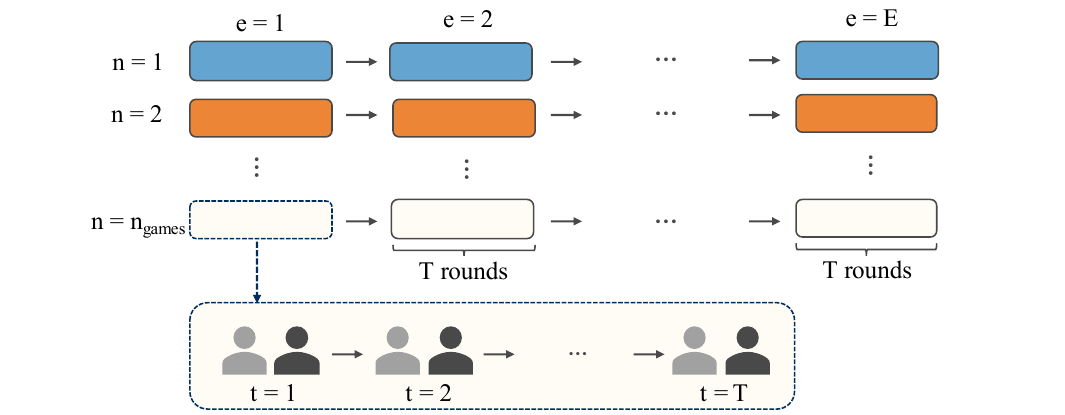}
\end{center}
\caption{Schematic representation of a trial. Each box corresponds to an episode (a game played for $T$ rounds). Same-colored boxes represent episodes within the same parallel environment. Within each environment, episodes occur sequentially as indicated by the arrows. The shaper updates its parameters using the experience collected throughout the entire trial.}
\label{trial_diagram}
\end{figure}

We introduce \textit{ShapeLLM}, a model-free opponent shaping algorithm designed to leverage the natural language capabilities of LLMs. ShapeLLM condenses both history and context into structured natural language prompts, explicitly capturing the two forms of memory required for shaping into one information stream. As in existing model-free algorithms \citep{lu2022mfos, khan2024shaper}, interactions are organized into trials. Each trial comprises $n_\text{games}$ parallel environments, where agents engage in $E$ episodes, each comprising $T$ rounds of the specified matrix game (Figure~\ref{trial_diagram} provides a schematic representation of a trial).

Let $t \in \{1, \ldots, T\}$ denote the round within an episode, and $e \in \{1, \ldots, E\}$ denote the episode within a trial. For notational simplicity, we collapse the pair $(e, t)$ into a single timescale index $\tau \in \{1, \ldots, E \times T\}$. We formalize the shaping task as a POMDP $(\bar{\mathcal{S}}, \bar{\mathcal{A}}, \bar{\mathcal{P}}, \bar{\mathcal{R}}, \bar{\Omega}, \bar{\mathcal{O}}, \bar{\gamma})$. The state $\bar{s}^\tau = \{ \theta_i^{\tau-1}, c_{i}^{\tau-1} \}_{i \in I} \in \bar{\mathcal{S}}$ encodes the parameters and conditioning prompts of all LLM agents from the previous timestep. The action space $\bar{\mathcal{A}}$ and reward function $\bar{\mathcal{R}}$ are equivalent to those of the underlying repeated normal-form game. The observation $\bar{o}^\tau = f(\mathbf{a}^1, \mathbf{a}^2, \ldots, \mathbf{a}^{\tau-1})$ is a function of the joint actions across all past timesteps. Lastly, $\bar{\mathcal{P}}$ and $\bar{\mathcal{O}}$ denote the state transition and observation functions, respectively.

At the beginning of training, players are initialized with policies $\{\rho_{\theta_i}^0\}_{i \in I}$ and receive initial observations $\{c^0_i\}_{i \in I}$ specifying the game characteristics (number of players, action space, and reward matrix) and action labels. At the $\tau$-th round of a trial, the shaper receives an observation $c^\tau_j$ that concatenates two components: the most recent joint action ($\mathbf{a}^{\tau-1}$) and a compressed natural language representation of all the previous joint actions in the trial ($f(\mathbf{a}^1, \ldots, \mathbf{a}^{\tau-2})$). This separation captures the distinction between history and context. The shaper then samples an action $a_j^{\tau} \sim \rho_{\theta_j}(w \mid c_j^{\tau})$ and receives the corresponding reward $r^\tau_j$ and next observation $c^{\tau+1}_j$.

Opponents update their policy parameters between episodes (i.e., when $\tau$ is a multiple of $T$) using the experience collected in the preceding episode. Consequently, within each trial, the shaper is exposed to $E$ opponent updates, though only indirectly through the evolving summaries of joint actions that persist across episodes. By contrast, the shaper’s own parameters $\theta_j$ are updated only at trial finalization to maximize the cumulative trial return $\bar{J} = \sum_{\tau=1}^{E \times T} r^\tau_j$. While we present this formulation in the context of repeated normal-form games, the ShapeLLM framework generalizes to any environment that can be formulated as a partially observable stochastic game.

\section{Experimental Settings}

\subsection{Environments}

We investigate opponent shaping on iterated versions of four canonical $2 \times 2$ games. These environments were selected to represent diverse incentive structures across strategic interactions.

\textbf{Iterated Prisoner's Dilemma (IPD)}. Players choose between cooperation (C) and defection (D). Mutual cooperation yields the highest collective payoff, but each player faces individual incentives to defect and exploit cooperative opponents \citep{rapoport1974pd, axelrod1981evolution}. 

\textbf{Iterated Matching Pennies (IMP)}. A zero-sum game where players choose between heads (H) and tails (T). One player receives a positive payoff when actions match, whereas the other is rewarded when they differ. This environment is purely adversarial. 

\textbf{Iterated Chicken Game (ICG)}. Players can either Swerve (S) or Go straight (G). Going straight yields an advantage against a swerving opponent, but mutual aggression results in catastrophic outcomes for both, creating a coordination problem under risk \citep{rapoport1966chicken}.

\textbf{Iterated Stag Hunt (ISH)}. Players choose between Stag (S) and Hare (H). Hunting stag yields the highest payoff but only if both players coordinate, while hunting hare offers a lower but guaranteed reward. This creates a coordination problem with multiple equilibria.

We assign a single token $w_{a_i}$ to each action $a_i \in \mathcal{A}$ (e.g. ``C" for cooperate and ``D" for defect in the IPD), and treat any other generation as an illegal action ($a_\text{null}$). Choosing $a_\text{null}$ incurs a penalty $r_\text{null}$, and the transition is excluded from both players' game histories  (see Appendix~\ref{app:LLM_game_imp}). 

\subsection{Implementation Details}
\label{sec:implementation_details}

Our base model is \textit{gemma-2-2b-it} \citep{gemma2024}, a small, instruction-tuned, open-source language model. We focus on small models for computational efficiency and choose instruction-tuned variants as they are more goal-directed \citep{ouyang2022training} and benefit from coding data exposure \citep{duan2024gtbench}. To keep the agent architecture minimal, we restrict memory to the context window of the model and avoid additional reasoning scaffolds such as chain-of-thought (CoT) prompting \citep{wei2022chain}, which are less effective in small models \citep{wei2022emergent}.

We train our agents using QLoRA \citep{dettmers2023qlora}, with the base model quantised to 4-bit precision via the \texttt{BitsAndBytes} package \citep{bitsandbytes2022}, and adapters of rank $r=2$ implemented through the \texttt{PEFT} library \citep{peft2022}. The learnable parameters comprise the LoRA adapters for the query/value projections and the value head parameters. All models are fine-tuned using a custom implementation of PPO that inherits from the \texttt{TRL} package\footnote{The default implementation is only compatible with contextual bandits.} \citep{trl2020}. We run PPO training for 200-300 epochs, with $n_\text{games}=5$ parallel environments, $E=5$ episodes, and $T=20$ rounds per episode. For the shapers, we express context via cumulative state visitation counts\footnote{We represent the context via visitation counts instead of full trajectories to prevent the token length from growing linearly with the number of rounds.} (e.g., in the IPD: ``CC: 1, CD: 1, DC: 2, DD: 3"). All training was done on a single A100 GPU with 40G of VRAM. The full specification of the hyperparameters, reward matrices, and training prompts used is provided in Appendices~\ref{app:training_hyperparams},~\ref{app:game_payoffs},~\ref{app:training_prompts}, respectively. 

\section{Shaping in Exploitative Settings}
\label{sec:shaping_exploitative}

We consider two core training configurations for our agents across the IPD, IMP and ICG. 

\textbf{Baseline}. We establish the baseline performance using two LLM-based \textit{naive learners} (NL) that treat their opponent as a stationary component of the environment. Each agent's conditioning prompt contains the game description and the most recent joint action. Once the episode is finished, both players simultaneously update their parameters via PPO to maximize episodic returns. This baseline establishes expected behavior when no opponent shaping occurs.

\textbf{Shaper vs. naive learner}. An LLM-based shaper interacts with an LLM naive learner (with the same configuration as in the baseline). As described in Section~\ref{sec:opp_shaping}, the shaper updates its parameters only at trial completion, after having observed multiple opponent parameter updates. 

Following training, we evaluate performance by having each trained pair of agents play 100 games with the same episode length used during training ($T=20$).

\subsection{Shaping in the IPD, IMP and ICG}
\label{sec:main_results}

\begin{figure}[h]
\begin{center}
\includegraphics{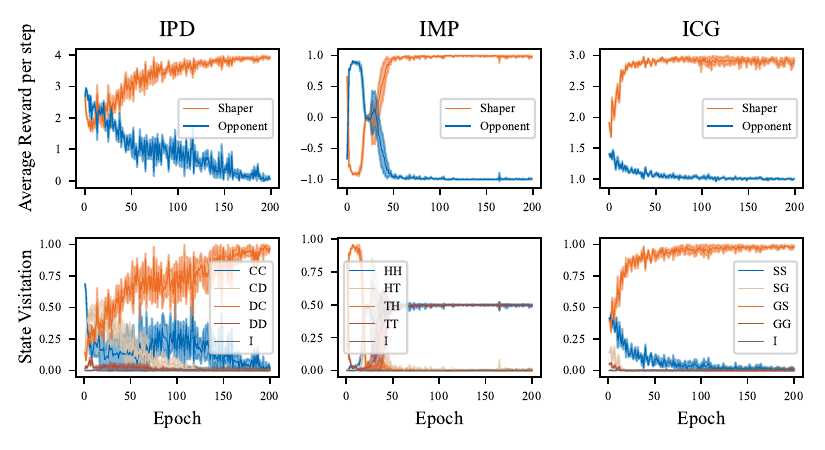}
\end{center}
\caption{Average reward per step (top row) and state visitation (bottom row) during training for the shaping experiments across the IPD, IMP, and ICG. In the state visitation figures, the outcome ``I" encompasses all transitions where either player chose $a_\text{null}$. Results are reported along with a 95\% confidence interval over 5 random seeds.}
\label{fig:main_results}
\end{figure}

We run experiments across 5 random seeds\footnote{For the ICG baseline, we conducted experiments over 10 seeds. Each seed converges to one of the two possible Nash Equilibria, so a larger sample size is needed to accurately estimate the percentage of convergence to each equilibrium.} using action labels $w_{a_1} = \text{C}, w_{a_2} = \text{D}$ for the IPD, $w_{a_1} = \text{H}, w_{a_2} = \text{T}$ for the IMP, and $w_{a_1} = \text{S}, w_{a_2} = \text{G}$ for the ICG. Figure~\ref{fig:main_results} illustrates the training dynamics for the shaper experiments across the three games (the corresponding figures for the baselines are shown in Appendix~\ref{app:training_dynamics}).

Table~\ref{tab:main_results} presents the post-training evaluation results, where each jointly-trained pair played 100 games with episode length $T=20$. The results demonstrate successful opponent shaping across all environments. For completeness, we evaluate performance across games of varying lengths (see Appendix~\ref{app:different_T}) and observe a similar performance.

\begin{table}[h]
\caption{Post-training evaluation results for the IPD, IMP, and ICG comparing baseline (two naive learners) versus shaper-naive learner pairs. Average rewards per step are reported with 95\% confidence intervals across 5 random seeds, except for the ICG baseline, where we use 10. Transitions involving $a_\text{null}$ are excluded (comprising 2\% of actions in IPD, 0.1\% in IMP, and 1\% in ICG).}
\label{tab:main_results}
\centering
\begin{tabular}{@{}lcccc@{}}
\toprule
\multirow{2}{*}{} & \multicolumn{2}{c}{\textbf{Baseline}} & \multicolumn{2}{c}{\textbf{One Shaper}} \\
\cmidrule(lr){2-3} \cmidrule(lr){4-5}
& Player 1 & Player 2 & Shaper & Opponent \\
\midrule
IPD & $1.00 \pm 0.00$ & $1.00 \pm 0.00$ & $3.96 \pm 0.01$ & $0.10 \pm 0.04$ \\
IMP & $-0.03 \pm 0.09$ & $0.03 \pm 0.09$ & $0.99 \pm 0.01$ & $-0.99 \pm 0.01$ \\
ICG & $2.00 \pm 0.58$ & $2.00 \pm 0.58$ & $2.98 \pm 0.01$ & $1.01 \pm 0.01$ \\
\bottomrule
\end{tabular}
\end{table}

We begin by examining the performance in the IPD. In the baseline, both learners converge to mutual defection, which is the Nash Equilibrium, achieving an average payoff of 1. In contrast, the shaper achieves an average reward of 3.96, exceeding what any zero-determinant extortion \citep{press2012iterated} or tit-for-tat strategy could obtain. Meanwhile, the opponent achieves 0.1, which is lower than the mutual defection payoff. The training dynamics show a three-phase pattern: starting from high initial cooperation, the shaper first sharply reduces its cooperation rate, then plateaus at a stable level to maintain opponent cooperation, and finally slowly decreases cooperation to achieve near-maximal exploitation. 

We observe similar performance patterns in the ICG. The baseline results show average rewards of 2 for both players. The standard deviations of 0.58 reflect the fact that each seed converges to one of the two pure Nash equilibria: either (Swerve, Go Straight) or (Go Straight, Swerve). In contrast, the shaper consistently achieves an average reward of 2.98 while limiting its opponent to 1.01. The training dynamics differ from the IPD: the shaper adopts an aggressive strategy by sharply reducing its swerving probability early in training, forcing convergence to its preferred equilibrium.

Finally, examining the IMP results, we observe that both agents in the baseline oscillate around the mixed Nash equilibrium with near-zero average payoffs (-0.03 and 0.03 respectively). In contrast, when one agent is a shaper, we observe clear exploitation, with the shaper obtaining a reward of 0.99 while the opponent obtains -0.99. The state visitation converges to equal frequency for the two states that favor the shaper: (H, H) and (T, T).

Collectively, these results demonstrate that the shaper consistently outperforms its opponent across all games, providing strong evidence of its ability to successfully influence opponents' learning dynamics in adversarial and mixed-motive scenarios. To establish the robustness of these results, we conduct two additional sets of experiments. First, we conduct an ablation study to determine whether shaping effects stem solely from the shaper's enriched observation space (Appendix~\ref{app:enriched_baseline}). We adopt the same settings as in the baseline experiments, with one of the agents receiving a summary of all interactions in the current episode. Second, we examine sensitivity to prompt variations by running shaping experiments with actions presented in reversed order and an alternative prompt formulation (Appendix~\ref{app:robustness_checks}). These experiments confirm that ShapeLLM achieves robust shaping across different configurations, and that enriched observations alone are insufficient for effective opponent shaping.

\subsection{Robustness against Different Opponents}
\label{sec:main_results_robust}
A robust shaping procedure should be capable of successfully influencing the learning dynamics of a diverse set of opponents. To test this capability, we explore shaping against opponents with distinct initial policies. For each game, we systematically select three action label pairs yielding initial probabilities of playing action $a_1$\footnote{The action $a_1$ corresponds to cooperation in the IPD, playing heads in the IMP, and swerving in the ICG.} of 0.75, 0.5, and 0.25. We then conduct shaping experiments using these labels for the opponents. A detailed description of the selection method, action labels, and initial output probabilities for each opponent is provided in Appendix~\ref{app:init_policies_opp}. We train shapers against each selected opponent using the same procedure as in Section~\ref{sec:main_results}, with evaluation results shown in Table~\ref{tab:diff_opp}.

\begin{table}[h]
\caption{Post-training evaluation results for the shaping experiments in the IPD, IMP and ICG across different opponent initializations. Each column represents a distinct opponent characterized by its approximate initial probability of playing action $a_1$. Average rewards per step are reported with 95\% confidence intervals across 5 seeds. Transitions where $a_\text{null}$ is played by either player are excluded from the analysis (comprising 0-2\% of all transitions).}
\label{tab:diff_opp}
\centering
\begin{tabular}{@{}lcccccc@{}}
\toprule
\multirow{2}{*}{} & \multicolumn{2}{c}{$p^0_{\text{NL}}(a_1) \sim 0.75$} & \multicolumn{2}{c}{$p^0_{\text{NL}}(a_1) \sim 0.50$} & \multicolumn{2}{c}{$p^0_{\text{NL}}(a_1) \sim 0.25$} \\
\cmidrule(lr){2-3} \cmidrule(lr){4-5} \cmidrule(lr){6-7}
& Shaper & Opponent & Shaper & Opponent & Shaper & Opponent \\
\midrule
IPD & $3.99 \pm 0.01$ & $0.01 \pm 0.02$ & $3.95 \pm 0.01$ & $0.04 \pm 0.03$ & $3.98 \pm 0.02$ & $0.07 \pm 0.07$ \\
IMP & $0.96 \pm 0.02$ & $-0.96 \pm 0.02$ & $0.99 \pm 0.01$ & $-0.99 \pm 0.01$ & $0.99 \pm 0.01$ & $-0.99 \pm 0.01$ \\
ICG & $3.00 \pm 0.00$ & $1.00 \pm 0.00$ & $2.99 \pm 0.01$ & $1.01 \pm 0.01$ & $2.95 \pm 0.01$ & $1.05 \pm 0.01$ \\
\bottomrule
\end{tabular}
\end{table}

Across all games and opponent types, shapers successfully exploit their co-players, achieving average per-timestep rewards of 3.97 in the IPD, 0.98 in the IMP, and 2.98 in the ICG. In the IPD and ICG, opponents converge to less favorable outcomes when initialized with more cooperative policies: their average rewards per step range from 0.01 to 0.07 in the IPD and from 1.00 to 1.05 in the ICG as initial policies become increasingly defective. The training dynamics (Appendix~\ref{app:training_dynamics}) reveal that shapers respond strategically to opponent initialization. Against more cooperative opponents, shapers reduce their own cooperation more rapidly and reach lower final cooperation levels. This pattern indicates that initially defective agents require more prolonged cooperation incentives before they can be effectively exploited.

In contrast, the IMP shows no sensitivity to different opponent initializations. Intuitively, shaping should be more challenging against initial policies closer to the mixed Nash equilibrium. However, our results reflect no such effect. It is worth noting that the opponent with $p^0_{\text{NL}} \sim 0.5$ is not initialized with a purely random policy, but with the closest approximation achievable through action label selection (see Appendix~\ref{app:init_policies_opp} for the exact initial policy). Shapers converge to near-optimal outcomes, achieving rewards of 0.96, 0.99, and 0.99.

\section{Shaping in Cooperative Settings}
\label{sec:shaping_cooperative}

\begin{figure}[ht]
\begin{minipage}{0.45\textwidth}
We investigate whether shaping can guide interactions toward mutually beneficial rather than purely exploitative outcomes. We explore this in two environments: a cooperative variation of the IPD (C-IPD) and the Iterated Stag Hunt (ISH). For the C-IPD, we provide both agents with the original payoff matrix, but modify the shaper's version so that its highest payoff is achieved through mutual cooperation, with all other payoffs unchanged\footnotemark. This could be interpreted as the shaper receiving an intrinsic reward when the most globally beneficial outcome is achieved. The reward matrices used can be found in Appendix~\ref{app:game_payoffs}. We use action labels $w_{a_1}=\text{C}, w_{a_2}=\text{D}$ for the C-IPD, and $w_{a_1}=\text{S}, w_{a_2}=\text{H}$ for the ISH. Figure~\ref{fig:coop_results} illustrates the training dynamics for the shaper experiments across the two environments (the corresponding figures for the baselines are shown in Appendix~\ref{app:training_dynamics}). Table~\ref{tab:coop_main_results} presents the post-training evaluation results. 
\end{minipage}\hfill
\begin{minipage}{0.5\textwidth}
\includegraphics[width=\textwidth]{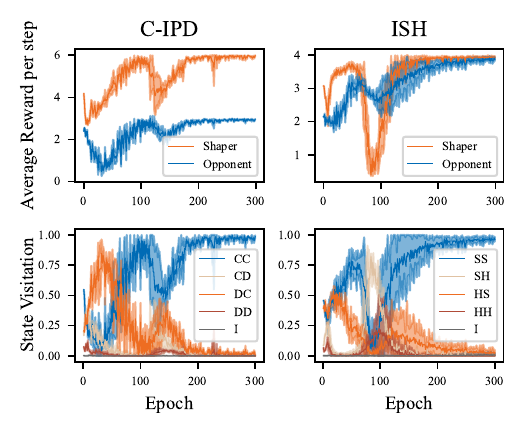}
\caption{Average reward per step (top row) and state visitation (bottom row) during training for the shaping experiments across the C-IPD and ISH.  In the state visitation figures, the outcome ``I" encompasses all transitions where either player chose $a_\text{null}$. All results are reported along with a 95\% confidence interval over 5 random seeds.}
\label{fig:coop_results}
\end{minipage}
\end{figure}

\footnotetext{We considered a variant where the shaper received the sum of both players' payoff as a reward. This configuration shifted the Nash equilibrium to an asymmetric outcome where the shaper cooperates while the opponent defects.}

In the ISH baselines both agents achieve a mean reward of 1.30, where 90\% of runs converge to the Pareto-inferior equilibrium (both hunt Hare)\footnote{We run the ISH and C-IPD baseline experiments across 10 seeds.}. With a shaper present, runs consistently converge to the Pareto-optimal equilibrium (both hunt Stag), both achieving rewards of approximately 3.96. This shows that shaping can resolve coordination failures and guide systems toward mutually beneficial outcomes when cooperation is required. In the cooperative IPD variant, baseline runs converge to the Nash equilibrium (mutual defection) with rewards of 1 each. With a shaper, all runs achieve mutual cooperation, yielding rewards of 5.88 and 2.86 for the shaper and naive learner, respectively. This outcome demonstrates that shaping can achieve globally beneficial outcomes in environments where other players have mixed incentives.

\begin{table}[!htpb]
\caption{Post-training evaluation results for the C-IPD and ISH comparing baseline versus shaper-naive learner pairs. Average rewards per step are reported with 95\% confidence intervals across 5 and 10 random seeds for shaping and baseline experiments, respectively. Transitions with $a_{\text{null}}$ are excluded from the analysis ($\sim2\%$ and $\sim0.1\%$ of actions respectively).}
\label{tab:coop_main_results}
\centering
\begin{tabular}{@{}lcccc@{}}
\toprule
\multirow{2}{*}{} & \multicolumn{2}{c}{\textbf{Baseline}} & \multicolumn{2}{c}{\textbf{One Shaper}} \\
\cmidrule(lr){2-3} \cmidrule(lr){4-5}
& Player 1 & Player 2 & Shaper & Opponent \\
\midrule
C-IPD & $1.00 \pm 0.00$ & $1.00 \pm 0.00$ & $5.88 \pm 0.03$ & $2.86 \pm 0.02$ \\
ISH & $1.30 \pm 0.52$ & $1.30 \pm 0.52$ & $3.96 \pm 0.02$ & $3.96 \pm 0.02$ \\
\bottomrule
\end{tabular}
\end{table}

\section{Discussion}

\textbf{Implications.} In this work, we have demonstrated that LLM-based agents can be susceptible to opponent shaping in both exploitative and cooperative game-theoretic settings. As LLM agents become increasingly deployed in real-world applications, they will inevitably interact with other agents, and potentially train on data acquired through these interactions (for example, in the case of continually learning LLMs). In such settings, our findings suggest that agents could be vulnerable to strategic exploitation by opponents with no knowledge or control over them. Conversely, the same mechanisms could be leveraged beneficially, enabling agents to guide interactions toward mutually beneficial outcomes regardless of the goals of their opponents.

\textbf{Limitations.} Our work has several limitations that suggest promising directions for future research. First, due to computational constraints, we have only evaluated our approach using a single small model (\textit{gemma-2-2b-it}). Future work could investigate whether shaping capabilities generalize to larger models and explore the relationship between model scale and shaping dynamics. For instance, whether smaller models are more vulnerable to these influences or whether larger models possess enhanced shaping capabilities. Second, our experiments instructed agents to select from a fixed set of action tokens. While this restriction made evaluation tractable, it limits the ways in which LLMs can influence one another. In practice, LLM agents can communicate through natural language, and expanding interactions beyond fixed tokens may substantially alter shaping dynamics. Even within the same game-theoretic settings, agents could employ language strategically, for example, by signaling intentions or negotiating before making a move. Future work could examine whether such natural language interaction strengthens, weakens, or qualitatively changes shaping outcomes. Finally, our study was restricted to $2 \times 2$ matrix games, where incentives are unambiguous and easily interpreted. Many real-world interactions, however, involve more nuanced or overlapping objectives, where cooperation and competition are not strictly binary. Exploring shaping in environments with richer payoff structures or multiple objectives would yield a deeper understanding of how these dynamics generalize to more realistic settings.

\section{Conclusion}

In this paper, we have investigated whether opponent shaping, a well-established technique in multi-agent reinforcement learning, extends to LLM-based agents. To the best of our knowledge, this is the first work to study opponent shaping with LLM agents. We proposed ShapeLLM, a model-free opponent shaping method for transformer-based agents, and demonstrated successful shaping in both exploitative and cooperative settings. In exploitative scenarios, LLM shapers influenced opponent learning in repeated games such as the IPD, IMP, and ICG, steering convergence toward outcomes that maximized their own payoff. In cooperative scenarios, shaping promoted coordination in settings like the ISH and a modified IPD, guiding agents toward mutually beneficial equilibria.  By demonstrating that LLMs can both shape and be shaped through interaction alone, our findings highlights the importance of understanding multi-agent dynamics when deploying these systems in shared environments.

\section*{Ethics Statement}

This work investigates opponent shaping in LLM agents. Shaping can promote coordination and prosocial behavior in multi-agent interactions. Even in mixed-motive scenarios,  it can avoid convergence to suboptimal outcomes, such as mutual defection in the IPD. Despite these advantages, shaping also poses inherent risks. These same techniques can be used for strategic exploitation, where an agent manipulates others' learning dynamics for unilateral advantage. As LLM agents are increasingly adopted in real-world environments, understanding these dynamics becomes crucial for responsible deployment. While our experiments are conducted in game-theoretic environments that do not pose immediate risks for real-world agent deployment, this work aims to raise awareness of both the opportunities and risks inherent in multi-agent LLM interactions, and it may serve as a basis for developing countermeasures against adversarial behavior.

\section*{Acknowledgments}

Marta Emili Garcia Segura was supported by the EPSRC through the Centre for Doctoral Training Studentship in Cybersecurity (EP/S022503/1). 

\bibliography{conference}
\bibliographystyle{conference}

\appendix
\section{Appendix}

\subsection{Payoff Matrices and illegal action penalties used during training}
\label{app:game_payoffs}

In Section~\ref{sec:shaping_exploitative}, we consider three environments: Iterated Prisoner's Dilemma (IPD), Iterated Matching Pennies (IMP), and the Iterated Chicken Game (ICG). Table~\ref{tab:payoffs} shows their corresponding payoff matrices, where each cell contains a tuple representing the payoffs of the row (first entry) and column (second entry) players. Actions are represented with the following labels: ``C" for cooperate and ``D" for defect in IPD, ``H" for heads and ``T" for tails in IMP, and ``S" for swerve and ``G" for go straight in ICG.

\begin{table}[h]
\caption{Payoff matrices for the three environments considered in Section~\ref{sec:shaping_exploitative} to explore shaping in exploitative settings.}
\label{tab:payoffs}
\centering
\begin{tabular}{ccc}
\begin{tabular}{c|cc}
\multicolumn{3}{c}{\bf (a) IPD} \\
& C & D \\
\hline
C & (3, 3) & (0, 4) \\
D & (4, 0) & (1, 1) \\
\end{tabular}
&
\begin{tabular}{c|cc}
\multicolumn{3}{c}{\bf (b) IMP} \\
 & H & T \\
\hline
H & (1, -1) & (-1, 1) \\
T & (-1, 1) & (1, -1) \\
\end{tabular}
&
\begin{tabular}{c|cc}
\multicolumn{3}{c}{\bf (c) ICG} \\
 & S & G \\
\hline
S & (2, 2) & (1, 3) \\
G & (3, 1) & (-5, -5) \\
\end{tabular}
\end{tabular}
\end{table}

In Section~\ref{sec:shaping_cooperative}, we use opponent shaping to promote globally beneficial outcomes in two games: Iterated Stag Hunt (ISH) and a cooperative IPD variant (C-IPD). In the latter, one player receives an enhanced reward for mutual cooperation while all other payoffs remain unchanged from the standard IPD. Table~\ref{tab:payoffs_coop} shows the corresponding payoff matrices,  where C-IPD uses the same action labels as IPD, and ISH uses ``S" for stag and ``H" for hare.

\begin{table}[h]
\caption{Payoff matrices for two environments considered in Section~\ref{sec:shaping_cooperative} to investigate shaping in cooperative settings.}
\label{tab:payoffs_coop}
\centering
\begin{tabular}{cc}
\begin{tabular}{c|cc}
\multicolumn{3}{c}{\bf (a) C-IPD} \\
& C & D \\
\hline
C & (6, 3) & (0, 4) \\
D & (4, 0) & (1, 1) \\
\end{tabular}
&
\begin{tabular}{c|cc}
\multicolumn{3}{c}{\bf (b) ISH} \\
 & S & H \\
\hline
S & (4, 4) & (0, 3) \\
H & (3, 0) & (1, 1) \\
\end{tabular}

\end{tabular}
\end{table}

The penalty for generating an illegal action, $r_\text{null}$, is always set to one unit below the lowest reward in each game's payoff matrix. Specifically: $r_\text{null}^{\text{IPD}} = r_\text{null}^{\text{C-IPD}} = -1$, $r_\text{null}^{\text{IMP}} = -2$, $r_\text{null}^{\text{ICG}} = -6$ and $r_\text{null}^{\text{ISH}} = -1$.


\subsection{LLM Gameplay in Repeated Normal-Form Games} 
\label{app:LLM_game_imp}

All the environments considered are 2$\times$2 repeated normal-form games. Since our agents are LLM-based, even after restricting the generation length to one token, their output space is much larger than the game's action space ($|\mathcal{V}| >> |\mathcal{A}| = 2 $).  

To tackle this space mismatch, one could try shrinking the model's output space via logit masking or rejection sampling. However, these interventions can alter the masked logits in unexpected ways or lead to increased computational time. Instead of actively ignoring or hiding parts of the output space, we define a mapping $\phi: \mathcal{V} \to \mathcal{A}$. 

Directly defining such a mapping would require distributing the entire vocabulary across two actions, resulting in semantically unrelated tokens being mapped to the same action. This would impose an unnecessary learning objective whereby agents must learn arbitrary semantic equivalences that are orthogonal to the underlying strategic objective. To avoid this, we introduce a null action $a_\text{null}$, such that $\mathcal{A}_i^\prime = \mathcal{A}_i \cup \{a_\text{null}\}$. This action is not meant to represent refusal to engage in the game, but rather failure to produce a reasonable answer.

We then define $\phi_i: \mathcal{V} \to \mathcal{A}_i^\prime$. This formulation is general and can accommodate open-ended generation if the mapping itself is another language model. For simplicity, we choose: 
\begin{equation}
    \phi_i (w)=
\begin{cases} 
a_1 & \text{if } w = w_{a_1}, \\
a_2 & \text{if } w = w_{a_2}, \\
a_\text{null} & \text{otherwise },
\end{cases}
\end{equation}
such that each action can be played with one specific token, while any other token is considered \textit{illegal}. The generation is steered towards this format via textual instructions (e.g.,\textit{ Reply with ``C" or ``D"}).

Augmenting the action space requires extending the payoff matrix. Table-\ref{tab:aug-payoff} shows the augmented payoff matrix for a general 2$\times$2 matrix game. The revised matrix is identical to the original one when both players play legal actions. If an agent plays $a_{\text{null}}$, it receives a penalty $r_\text{null}$, regardless of its opponent's move. If the agent plays a legal action but its opponent does not, the transition is discarded.

\begin{table}[h]
\caption{Augmented payoff matrix for training LLM-agents in repeated normal-form games. When both agents play legal actions, payoffs match those of the underlying game. When an agent plays an illegal action, it receives a penalty $r_\text{null}$, regardless of the opponent's action. However, when an agent plays a legal action but the opponent plays an illegal action, the transition is discarded as it provides no meaningful learning signal (indicated by dashes in the matrix).}
\label{tab:aug-payoff}
\centering
\begin{tabular}{l|lll}
& \multicolumn{1}{c}{\bfseries $\mathbf{a_1}$} & \multicolumn{1}{c}{\bfseries $\mathbf{a_2}$} & \multicolumn{1}{c}{\bfseries $\mathbf{a_\text{null}}$} \\
\hline
$\mathbf{a_1}$ & $r(a_1, a_1)$ & $r(a_1, a_2)$ & \textcolor{burntorange}{--} \\
$\mathbf{a_2}$ & $r(a_2, a_1)$ & $r(a_2, a_2)$ & \textcolor{burntorange}{--} \\
$\mathbf{a_\text{null}}$ & \textcolor{darkblue}{$r_\text{null}$} & \textcolor{darkblue}{$r_\text{null}$} & \textcolor{darkblue}{$r_\text{null}$} \\
\end{tabular}
\end{table}

With the augmented action space and payoff structure defined, we can now describe how agents interact within this framework. In the $t$-th round, each agent receives a context $c^t_i$, which is a sequence of tokens consisting of the game description and a summary of the previous rounds of the game. A single token is then sampled from the output distribution of each agent ($w^t_i \sim \rho_{\theta_i} ( w| c^t_i)$) and subsequently mapped to a game action, such that $ \mathbf{a}^t = \{a^t_i = \phi_i (w_i^t)\}_{i \in I}$. The environment then returns the corresponding rewards ($\{r_i^t  = R_i (\mathbf{a}^t)\}_{i \in I}$) and contexts for the next round ($\{c^{t+1}_i = f (c^t_i, \mathbf{a}^t)\}_{i \in I}$).


\subsection{Training Implementation details for reproducibility}
\label{app:training_hyperparams}

This section provides detailed hyperparameter specifications that supplement the implementation details in Section~\ref{sec:implementation_details}. 

\textbf{Generation Parameters}. We use the same generation parameters for all agents across the training and evaluation phases. The configuration is kept to the default values with three exceptions: we enable sampling (\texttt{do\_sample=True}), disable top-$k$ generation (\texttt{top\_k=0}), and restrict the generation length to one token (\texttt{max\_new\_tokens=1}). All other parameters (e.g., \texttt{temperature=1.0}, \texttt{top\_p=1.0}) remain at default values.

\textbf{Adapter Configuration.} We employ the same adapter configuration for all agents. We only modify the rank parameter (\texttt{r=2}) to reduce compute. All other parameters (e.g., \texttt{lora\_alpha=32}, \texttt{lora\_dropout=0.05}, \texttt{target\_modules = ["q\_proj", "v\_proj"]}) are kept at their default values. 

\subsubsection{Naive Learner Hyperparameters}

We aimed to maintain hyperparameters at their default values for naive learners to simulate realistic scenarios where opponents cannot be controlled. However, several adjustments were necessary due to memory constraints and training instability. To reduce the memory consumption and compute, we reduced the adapter rank, batch size, and mini-batch size. These parameters were kept identical for both shapers and naive learners to ensure a fair comparison. Additionally, we observed that under the default settings some agents learned to generate a substantial amount of illegal actions. To counter this instability, we reduced the learning rate, the number of optimization epochs per batch (PPO epochs), and incorporated reward scaling to maintain stable training dynamics across all agents.

\begin{table}[h]
\caption{Naive learner hyperparameters used in all experiments. The \textit{``(default)"} flag indicates hyperparameters taking the default value in the \texttt{TRL} package.}
\label{tab:NL_params_for_all}
\begin{center}
\begin{tabular}{ll}
\toprule
\multicolumn{1}{c}{\bf Parameter} & \multicolumn{1}{c}{\bf Value} \\
\midrule
LoRA rank & 2 \\
LoRA target modules &  [``q\_proj", ``v\_proj"] (default)\\
Learning rate & $1.41 \times 10^{-6}$\\
Use adaptive KL control & Yes (default) \\ 
Starting KL coefficient & $0.2$ (default) \\ 
Target KL value & $6.0$ (default) \\ 
Horizon for adaptive KL control & $10000$ (default) \\ 
GAE $\gamma$ & $1.0$ (default) \\ 
GAE $\lambda$ & $0.95$ (default) \\ 
Clipping range & $0.2$ (default) \\ 
Value Function clipping & $0.2$ (default) \\ 
Value Function Loss coefficient & $0.2$ (default)\\ 
Batch Size & $100$  \\ 
Mini Batch Size & $10$  \\ 
Gradient Accumulation Steps & $1$ (default) \\ 
PPO epochs & $1$ \\ 
Score normalization & No (default) \\ 
Score scaling & Yes\\ 
\bottomrule
\end{tabular}
\end{center}
\end{table}

Table~\ref{tab:NL_params_for_all} presents the hyperparameters used for training naive learners across all experiments\footnote{These parameters were used for all naive learners and games except IMP. Under these parameters, naive learners playing the IMP converged to deterministic strategies rather than the expected mixed Nash equilibrium. We therefore reduced the learning rate to $1.41 \times 10^{-7}$ and the value function coefficient to $0.05$ for all IMP naive learners.}. Parameters marked with the \textit{``(default)"} flag indicate values that remained unchanged from the \texttt{TRL} library's default configuration.

\subsubsection{Shaper Hyperparameters}

For shapers, the hyperparameters used varied across games and opponents. However, several core parameters were held constant across all experiments, which are presented in Table~\ref{tab:shaper_params_for_all}. Where parameters deviate from default values, modifications were made either for computational efficiency (rank, batch size, mini batch size) or improved training stability (score scaling, PPO epochs)

\begin{table}[h]
\caption{Shaper hyperparameters fixed across all experiments. The \textit{``(default)"} flag indicates hyperparameters taking the default value in the \texttt{TRL} package.}
\label{tab:shaper_params_for_all}
\begin{center}
\begin{tabular}{ll}
\toprule
\multicolumn{1}{c}{\bf Parameter} & \multicolumn{1}{c}{\bf Value} \\
\midrule
LoRA rank & 2 \\
LoRA target modules &  [``q\_proj", ``v\_proj"] (default)\\
Use adaptive KL control & Yes (default) \\ 
Starting KL coefficient & $0.2$ (default) \\ 
Target KL value & $6.0$ (default) \\ 
Horizon for adaptive KL control & $10000$ (default) \\ 
GAE $\gamma$ & $1.0$ (default) \\ 
GAE $\lambda$ & $0.95$ (default) \\ 
Value Function clipping & $0.2$ (default) \\ 
Batch Size & $100$  \\ 
Mini Batch Size & $10$  \\ 
Gradient Accumulation Steps & $1$ (default) \\ 
PPO epochs & $1$ \\ 
Score normalization & No (default) \\ 
Score scaling & Yes\\ 
\bottomrule
\end{tabular}
\end{center}
\end{table}

We varied three main hyperparameters across experiments: the learning rate ($\textit{lr}$), the value function coefficient ($c_\text{VF}$) and the clipping range ($\epsilon_\text{p}$). The learning rate and clipping range were reduced mainly to increase stability during training. Under the default settings, we observed high variation in convergence outcomes across different random seeds. The value function coefficient $c_\text{VF}$ weights the value function term in the PPO loss. Shapers operate over much longer horizons than naive learners, requiring them to predict expected returns at the trial level rather than the episode level. This creates a challenging value prediction problem: the value function must estimate returns ranging from entire trials (for initial states) to single immediate rewards (for final states). Since the shaper's value head is randomly initialized, under the default $c_\text{VF}$ value, the initial value loss dominates the policy loss by orders of magnitude, leading to large gradients and causing training instability. A particularly problematic consequence is that agents sometimes learn to generate illegal tokens simply to make the value prediction task easier. We address this issue by reducing the value function coefficient to better balance the relative contributions of the value and policy losses.

\begin{table}[h]
\caption{Shaper's learning rate ($\textit{lr}$), value function coefficient ($c_\text{VF}$), and clipping range ($\epsilon_\text{p}$) for the experiments in Sections~\ref{sec:main_results}, ~\ref{sec:main_results_robust}, ~\ref{sec:shaping_cooperative}.}
\label{tab:shaper_hyperparameters}
\begin{center}
\begin{tabular}{llccc}
\toprule
& \bf Experiment & \bf \textit{lr} & \bf $\mathbf{c}_\text{VF}$ & \bf $\mathbf{\epsilon}_\text{p}$ \\
\midrule
\multirow{3}{*}{Section~\ref{sec:main_results}} 
& IPD & $1.41 \times 10^{-7}$ & $10^{-3}$ & $10^{-4}$ \\
& IMP & $3.41 \times 10^{-7}$ & $10^{-3}$ & $2 \times 10^{-1}$ \\
& ICG & $1.41 \times 10^{-7}$ & $10^{-3}$ & $2 \times 10^{-1}$ \\
\midrule
\multirow{9}{*}{Section~\ref{sec:main_results_robust}} 
&IPD with $p^0_{\text{NL}} (a_1) \sim 0.75$ & 
$1.41 \times 10^{-7}$ & $10^{-3}$ & $2 \times 10^{-1}$ \\
&IPD with $p^0_{\text{NL}} (a_1) \sim 0.5$  & 
$1.41 \times 10^{-7}$ & $5 \times 10^{-4}$ & $5 \times 10^{-3}$ \\
&IPD with $p^0_{\text{NL}} (a_1) \sim 0.25$  & 
$1.41 \times 10^{-7}$ & $3 \times 10^{-3}$ & $10^{-4}$ \\
&IMP with $p^0_{\text{NL}} (a_1) \sim 0.75$  & 
$4.41 \times 10^{-7}$ & $10^{-3}$ & $ 2 \times 10^{-1}$ \\
&IMP with $p^0_{\text{NL}} (a_1) \sim 0.5$  & 
$4.41 \times 10^{-7}$ & $10^{-3}$ & $ 2 \times 10^{-1}$ \\
&IMP with $p^0_{\text{NL}} (a_1) \sim 0.25$  & 
$6.41 \times 10^{-7}$ & $10^{-3}$ & $ 2 \times 10^{-1}$ \\
&ICG with $p^0_{\text{NL}} (a_1) \sim 0.75$  & 
$1.41 \times 10^{-7}$ & $10^{-3}$ & $2 \times 10^{-1}$ \\
&ICG with $p^0_{\text{NL}} (a_1) \sim 0.5$  &
$1.41 \times 10^{-7}$ & $10^{-3}$ & $ 2 \times 10^{-1}$ \\
&ICG with $p^0_{\text{NL}} (a_1) \sim 0.25$  & 
$6.41 \times 10^{-8}$ & $10^{-3}$ & $ 2 \times 10^{-1}$ \\
\midrule
\multirow{2}{*}{Section~\ref{sec:shaping_cooperative}} 
& C-IPD & $8.41 \times 10^{-8}$ & $5 \times 10^{-5}$ & $2 \times 10^{-1}$ \\
& ISH & $8.41 \times 10^{-8}$ & $10^{-3}$ & $2 \times 10^{-1}$ \\
\bottomrule
\end{tabular}
\end{center}
\end{table}

Table~\ref{tab:shaper_hyperparameters} shows the learning rate, value function coefficient, and clipping range used to train the shapers across all experiments in the main text. The remaining hyperparameters used during training for all experiments are shown in Table~\ref{tab:shaper_params_for_all}.

\subsubsection{Package Versions}

We used the following versions for the main Python packages:
\begin{itemize}
    \item \texttt{BitsAndBytes}: v0.45.0
    \item \texttt{PEFT}: v0.14.0
    \item \texttt{torch}: v2.5.1
    \item \texttt{transformers}: v4.47.0
    \item \texttt{TRL}: v0.11.4
\end{itemize}


\subsection{Ablation study: Naive Learners with Enriched Observations}
\label{app:enriched_baseline}

We conduct a variation of the baseline experiments to test whether the shaping effects we observe are solely a product of the shaper’s enriched observation space. The experimental setup involves two naive learners, with one player's observation space augmented to include the state counts from all previous interactions in the current episode. Crucially, these observations are reset at episode boundaries (unlike the shaper, whose observations persist across episodes within a trial). To ensure comparability, we use the same action labels as in Section~\ref{sec:main_results}: $w_{a_1}=\text{C}, w_{a_2}=\text{D}$ for the IPD, $w_{a_1}=\text{H}, w_{a_2}=\text{T}$ for the IMP, and $w_{a_1}=\text{S}, w_{a_2}=\text{G}$ for the ICG. Both learners use the hyperparameters presented in Table~\ref{tab:NL_params_for_all}.

\begin{table}[!htbp]
\caption{Post-training evaluation results for baselines where Player~1 receives enriched observations. Columns \textit{ICG} and \textit{ICG (alt. opp.)} correspond to experiments with varying action labels for Player 2 (\textit{ICG} uses $w_{a_1} =\text{S}, w_{a_2}=\text{G}$, and \textit{ICG (alt. opp.)} employs $w_{a_1}=\text{N}, w_{a_2}=\text{M}$). Average rewards per step are reported with 95\% confidence intervals across 5 random seeds.}
\label{tab:enriched_base}
\begin{center}
\begin{tabular}{lcccc}
\toprule
\rule{0pt}{2.ex}& \bf IPD &\bf  IMP & \bf ICG & \bf ICG (alt. opp.)\\
\midrule
\rule{0pt}{2.ex}Player 1 (enriched) & $1.00 \pm 0.00$ & $-0.05 \pm 0.09$ & $2.60 \pm 0.76$ & $1.00 \pm 0.00$ \\ 
Player 2 & $1.00 \pm 0.00$ & $0.05 \pm 0.09$ & $1.38 \pm 0.74$ & $3.00 \pm 0.00$ \\
\bottomrule
\end{tabular}
\end{center}
\end{table}

Table~\ref{tab:enriched_base} reports the post-training average reward per step for the IPD, IMP, and ICG. In the IPD and IMP, despite the asymmetry in observations, outcomes mirror those in the standard baseline: mutual defection (1 each) for the IPD, and oscillation around the Nash Equilibrium (-0.05 and 0.05) for the IMP. This equivalence to the baseline is not the same for the ICG. When both players use action labels $w_{a_1}=\text{S}, w_{a_2}=\text{G}$ (row \textit{ICG}), the player with enriched observations achieves 2.60, while its opponent receives only 1.38. As in the baseline, each run converges to one of the two pure Nash equilibria, but 80\% of runs favor the equilibrium most beneficial to the player with enriched observations. At first glance, this suggests that the additional information, rather than training with ShapeLLM, could drive the shaping behavior observed in the ICG.

\begin{figure}[h]
\begin{center}
\includegraphics{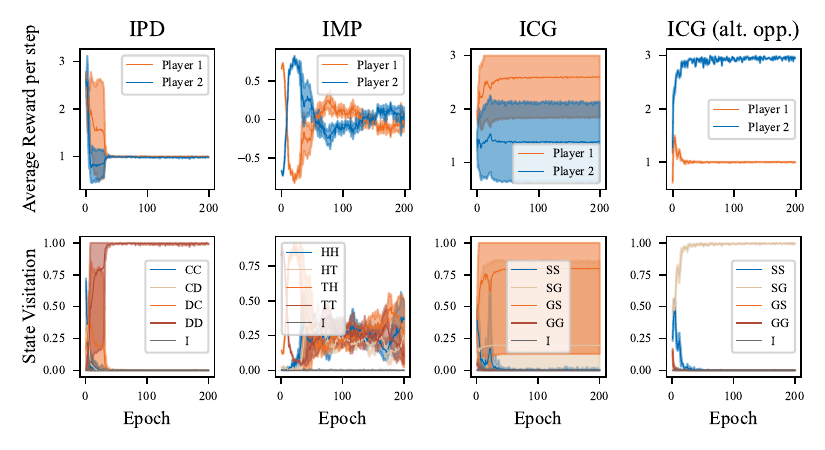}
\end{center}
\caption{Average reward per step (top row) and state visitation (bottom row) during training for the enriched observation baseline experiments across the IPD, IMP, and ICG. For the latter, two opponent configurations are presented: \textit{ICG} and \textit{ICG (alt. opp.)}. They use $w_{a_1}=\text{S}, w_{a_2}=\text{G}$ and $w_{a_1}=\text{N}, w_{a_2}=\text{M}$ as the opponent's action labels respectively, and $w_{a_1}=\text{S}, w_{a_2}=\text{G}$ for the player with enriched observations. In the state visitation figures, the outcome ``I”
encompasses all transitions where either player
chose $a_\text{null}$. The results are reported along with a 95\% confidence interval over 5 random seeds.}
\label{fig:enriched_baseline_plots}
\end{figure}

Inspecting the training dynamics (Figure~\ref{fig:enriched_baseline_plots}), we observe a skewed initialization in the ICG, with 40\% of transitions being (Go Straight, Swerve) at training initiation. To test whether the obtained results were a  consequence of this initialization, we conduct a control experiment where Player~2 selects actions with labels $w_{a_1}=\text{N}, w_{a_2}=\text{M}$\footnote{These labels are used in Section~\ref{sec:main_results_robust} to show the shaping robustness under different opponent initializations.}. For this setup, the initial state visitation is starkly different (see Figure~\ref{fig:enriched_baseline_plots}), with  46\% of transitions being (Swerve, Go Straight). In this case, the enriched baseline for the ICG consistently converges to the least favorable equilibrium for Player~1 (with rewards of 1.00, 3.00 for Players~1 and 2, respectively).

This control shows that the apparent shaping advantage in the ICG enriched baseline was driven by favorable initialization, not by the enriched observation space itself. When the initialization is modified, the enriched learner systematically converges to the least favorable outcome. In contrast, when ShapeLLM is applied to the same opponent configuration, the shaper consistently achieves maximum rewards (see Section~\ref{sec:main_results_robust}). Taken together, these results demonstrate that enriched observations alone are insufficient to produce shaping. Shaping requires the ability to indirectly observe and respond to the opponent's learning dynamics across episodes.

\subsection{Robustness to Prompt Variations in Exploitative Shaping Experiments}
\label{app:robustness_checks}

We conduct two experiments for robustness to prompt variations. The first variation uses a prompt where the payoff matrix is presented in table form. 

\begin{figure}[h]
\begin{center}
\begin{tcolorbox}[
    colback=lightblue,
    colframe=darkgray,
    fonttitle=\bfseries\color{white},
    fontupper=\ttfamily\fontsize{8pt}{10pt}\selectfont, 
    title={\textit{Table-format base prompt} for the IPD with $w_{a_1} = \text{C}, w_{a_2} = \text{D}$},
    coltitle=white,
    colbacktitle=darkblue,
    rounded corners,
    boxrule=2pt,
    width=\textwidth 
]
\texttt{<bos><start\_of\_turn>user}

\texttt{You are playing a 2-player game with actions: C, D. Points are assigned as follows:}

\begin{verbatim}

|       |  **C**  |  **D**  |
|-------|---------|---------|
| **C** |  (3, 3) |  (0, 4) |
| **D** |  (4, 0) |  (1, 1) |

\end{verbatim}

\texttt{Choose an action for the current round. Reply only with C or D.}

\texttt{<end\_of\_turn>}

\texttt{<start\_of\_turn>model}
\end{tcolorbox}
\end{center}
\caption{Table-format prompt variation for the IPD. Instead of a textual description, the payoff matrix is presented in markdown table form, replicating the base model's formatting style.}
\label{fig:table_prompt}
\end{figure}

To determine the specific table formatting, we ask the base model to generate a general payoff matrix and replicate its output format (including spacing and header formatting). Figure~\ref{fig:table_prompt} shows the \textit{table-format base prompt} used for the IPD.

\begin{table}{h}
\centering
\caption{Post-training evaluation results for the IPD, IMP, and ICG with the table-format and switched-order prompts. Average rewards per step are reported with 95\% confidence intervals across 5 random seeds. Transitions where $a_\text{null}$ is played by either player are excluded
from the analysis (comprising 0-4\% of all transitions)}
\vspace{0.5em}
\label{tab:evaluation_varying_prompts}
\begin{tabular}{lcccc}
\toprule
\rule{0pt}{2.ex}& \multicolumn{2}{c}{\bf Table-format} & \multicolumn{2}{c}{\bf Switched-order}\\
\cmidrule(lr){2-3} \cmidrule(lr){4-5}
& Shaper & Opponent & Shaper & Opponent \\
\midrule
IPD & $3.50 \pm 0.16$ & $0.53 \pm 0.09 $ & $3.99 \pm 0.01$ & $0.02\pm 0.01$\\
IMP & $0.94 \pm 0.06$ & $-0.94 \pm 0.06$ & $0.98\pm 0.04$ & $-0.98\pm 0.04$\\
ICG & $2.99 \pm 0.01$ & $1.01 \pm 0.01$ & $2.77 \pm 0.37$ & $1.23 \pm 0.37$\\
\bottomrule
\end{tabular}
\end{table}

\begin{figure}[h]
\begin{center}
\includegraphics{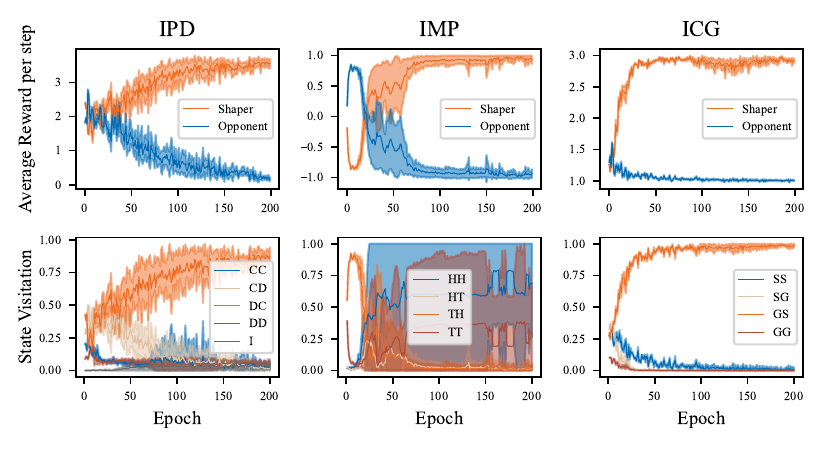}
\end{center}
\caption{Average reward per step (top row) and state visitation (bottom row) during training for the shaping experiments with the table-format prompt across the IPD, IMP, and ICG. In the state visitation figures, the outcome ``I" encompasses all transitions where either player
chose $a_\text{null}$. Results are reported along with a 95\% confidence interval over 5 random seeds.}
\label{fig:table_results}
\end{figure}

For the IMP and ICG, the prompts are identical except for the modified action labels and payoff matrices. The example in Figure~\ref{fig:table_prompt} shows the base prompt that both players receive at training initiation. The dynamic updates of this prompt throughout training (i.e., how history and context are updated) remain unchanged from those used in the main experiments (see Appendix~\ref{app:training_prompts}).  We use the same action labels as in Section~\ref{sec:main_results}: $w_{a_1}=\text{C}, w_{a_2}=\text{D}$ for the IPD, $w_{a_1}=\text{H}, w_{a_2}=\text{T}$ for the IMP, and $w_{a_1}=\text{S}, w_{a_2}=\text{G}$ for the ICG.  Figure~\ref{fig:table_results}, shows the training dynamics for the three games under the tabular prompt variation. It is worth noting that with this new prompt, both the initialization of the shaper and its opponent change significantly, especially for the IPD. Table~\ref{tab:evaluation_varying_prompts} shows the evaluation results obtained.

The results demonstrate successful shaping across all three games. In the IPD, the shaper achieves 3.5 while its opponent receives 0.53, exceeding mutual cooperation payoffs but underperforming compared to the main text results. Since Section~\ref{sec:main_results_robust} shows consistent outcomes across different opponent initializations, this gap likely stems from the shaper's own initialization rather than opponent effects. The IMP shows a similar pattern with substantial initialization differences, with the shaper achieving 0.94 versus the opponent's -0.94. For the ICG, where initialization more closely matches the main setup, results are 2.99 for the shaper and 1.01 for the opponent.

For the second prompt variation, we reverse the order in which actions are presented (see Figure~\ref{fig:diff_order_prompt}). As with the table-format experiments, the base prompts for IMP and ICG are identical except for their respective action labels and payoff matrices. We use the same action labels as in the previous variation.

\begin{figure}[h]
\begin{center}
\begin{tcolorbox}[
    colback=lightblue,
    colframe=darkgray,
    fonttitle=\bfseries\color{white},
    fontupper=\ttfamily\fontsize{8pt}{10pt}\selectfont, 
    title={\textit{Switched-order base prompt} for the IPD with $w_{a_1} = \text{C}, w_{a_2} = \text{D}$},
    coltitle=white,
    colbacktitle=darkblue,
    rounded corners,
    boxrule=2pt,
    width=\textwidth 
]
\texttt{<bos><start\_of\_turn>user}

\texttt{You are playing a 2-player game with actions:  D, C. Points are assigned as follows: D/D: 1/1,  D/C: 4/0, C/D: 0/4, C/C: 3/3.}

\texttt{Choose an action for the current round. Reply only with C or D.}

\texttt{<end\_of\_turn>}

\texttt{<start\_of\_turn>model}
\end{tcolorbox}
\end{center}
\caption{Switched-order prompt variation for the IPD. The actions and payoffs are presented in reversed order compared to the main text prompt.}
\label{fig:diff_order_prompt}
\end{figure}

\begin{figure}[h]
\begin{center}
\includegraphics{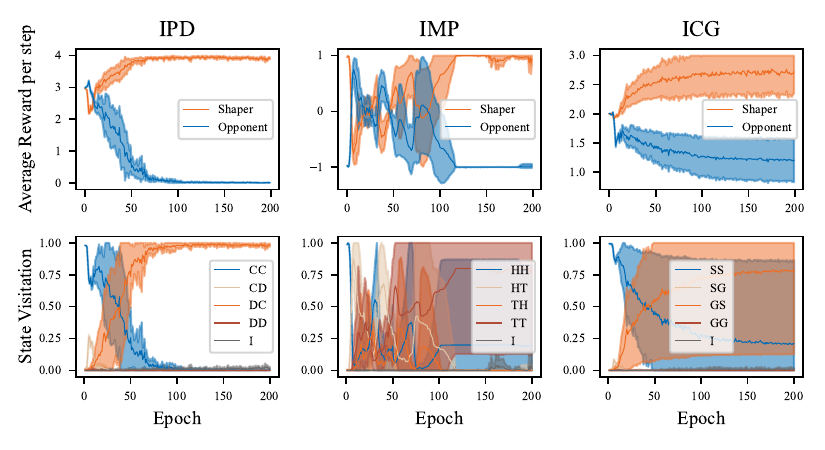}
\end{center}
\caption{Average reward per step (top row) and state visitation (bottom row) during training for the shaping experiments with the switched-order prompt across the IPD, IMP, and ICG. In the state visitation figures, the outcome ``I" encompasses all transitions where either player
chose $a_\text{null}$. Results are reported along with a 95\% confidence interval over 5 random seeds.}
\label{fig:diff_order_results}
\end{figure}

Figure~\ref{fig:diff_order_results} shows the training dynamics for the switched-order prompt experiments. As with the table-format variation, both agents exhibit substantially different initializations compared to the main experiments. Here, both players are heavily biased toward playing action $a_1$ (cooperate in the IPD, heads in the IMP, and swerving in the ICG), with their policies initially being almost deterministic (average of 99\% of ($a_1, a_1$) state at trial initiation). This creates challenges for the IPD and ICG, where this joint outcome yields acceptable rewards for both players. Consequently, without sufficient exploration incentives, the agents' policies remain unchanged for some of the seeds, focusing exclusively on the value prediction problem. 

To address this issue, we introduce an entropy regularization term to the PPO loss function. While this term was included in the original PPO formulation \citep{schulman2017proximal}, it is not implemented in the \texttt{TRL} package. Since our agents' action space encompasses the entire vocabulary, maximizing their output distribution entropy would create conflicting incentives: illegal actions would incur penalties while simultaneously reducing the loss through increased entropy. Instead, we extract logits only for the two allowed action tokens and compute the entropy of the resulting normalized distribution. To prevent convergence to suboptimal policies, we employ a decaying entropy coefficient that gradually reduces the exploration incentive during training.

The switched-order prompt results are presented in Table~\ref{tab:evaluation_varying_prompts}. With entropy regularization applied to both the IPD and ICG, we again observe successful shaping with results closely matching those in the main text: 3.99 for IPD and 2.77 for ICG. In the ICG, while the shaper achieves an average reward of 3 across most runs, for one of the seeds it consistently fails to explore alternative actions, converging to the outcome where both players swerve. Lastly, the IMP performs similarly to other prompt formulations (0.94 for the shaper vs -0.94 for its opponent) without requiring an entropy regularization term.

These experiments demonstrate that ShapeLLM maintains robust shaping capabilities across different prompt formulations. Even when initial policies are nearly deterministic (as in the switched-order variation), introducing entropy regularization to encourage early exploration enables effective opponent shaping. For reproducibility, we list the hyperparameters modified in this appendix in Table~\ref{tab:hyperparams_robust_appendix}. The rest of the hyperparameters used are specified in Table~\ref{tab:shaper_hyperparameters}. 

\begin{table}[h]
\caption{Shaper's learning rate ($\textit{lr}$), value function coefficient ($c_\text{VF}$), clipping range ($\epsilon_\text{p}$), and entropy regularization parameters ($c_{\text{S}}^\text{init}$, $c_{\text{S}}^\text{end}$, $T_\text{S}$) for the experiments in Appendix~\ref{app:robustness_checks}.}
\label{tab:hyperparams_robust_appendix}
\begin{center}
\begin{tabular}{llcccccc}
\toprule
& \bf Experiment & \bf \textit{lr} & \bf $\mathbf{c}_\text{VF}$ & \bf $\mathbf{\epsilon}_\text{p}$ &  $\mathbf{c}_{\text{S}}^\text{init}$ & $\mathbf{c}_{\text{S}}^\text{end}$ & $\mathbf{T}_\text{S}$\\
\midrule
\multirow{3}{*}{Table-format prompt} 
& IPD & $1.41 \times 10^{-7}$ & $10^{-3}$ & $10^{-5}$ & -- & -- & --\\
& IMP & $2.41 \times 10^{-7}$ & $10^{-3}$ & $2 \times 10^{-1}$ & -- & -- & --\\
& ICG & $1.41 \times 10^{-7}$ & $10^{-3}$ & $2 \times 10^{-1}$ & -- & -- & --\\
\midrule
\multirow{3}{*}{Switched-order prompt} 
& IPD & $6.41 \times 10^{-7}$ & $10^{-3}$ & $2 \times 10^{-1}$ & 0.1 & 0.0 & 25\\
& IMP & $6.41 \times 10^{-6}$ & $10^{-3}$ & $2 \times 10^{-1}$ & -- & -- & --\\
& ICG & $1.41 \times 10^{-7}$ & $10^{-3}$ & $2 \times 10^{-1}$ & 0.7 & 0.0 & 25\\
\bottomrule
\end{tabular}
\end{center}
\end{table}


\subsection{Action Label Selection for Robustness Experiments}
\label{app:init_policies_opp}

LLM agent output distributions vary significantly with the choice of action labels ($w_{a_1}, w_{a_2}$). For the robustness experiments, we systematically selected action labels to achieve target initial output distributions for the opponents.

\begin{table}[!htbp]
\caption{Action labels that produce the closest output distributions to the target initial distributions for the IPD, IMP and ICG.}
\label{tab:action_labels}
\begin{center}
\begin{tabular}{lcccccc}
\toprule
& \multicolumn{2}{c}{\bf$ \mathbf{p^0_{\text{NL}} (a_1) \sim 0.75}$} & \multicolumn{2}{c}{\bf$ \mathbf{p^0_{\text{NL}} (a_1) \sim 0. 50}$} & \multicolumn{2}{c}{\bf$  \mathbf{p^0_{\text{NL}} (a_1) \sim 0.25}$}\\
\cmidrule(lr){2-3} \cmidrule(lr){4-5} \cmidrule(lr){6-7}
&$w_{a_1}$ & $w_{a_2}$ & $w_{a_1}$ & $w_{a_2}$ & $w_{a_1}$ & $w_{a_2}$\\
\midrule
IPD& H & K & N & Y & I & X\\
IMP & S & Y & N & Y & N & M\\
ICG & T & K & N & M & T & F\\
\bottomrule
\end{tabular}
\end{center}
\end{table}

\textbf{Target Distributions}. For each game, we want to obtain opponents with initial probabilities of playing action $a_1$\footnote{The action $a_1$ corresponds to cooperation in the IPD, playing heads in the IMP, and swerving in the chicken game.} of approximately 0.75, 0.5, and 0.25. 

\textbf{Selection Procedure}. Opponents encounter 5 different prompts during training\footnote{One stateless prompt at training initiation, and 4 prompts corresponding to the 4 possible joint actions from the previous round (e.g., in the IPD: CC, CD, DC, DD).}. We extract the LLM's output probabilities for these 5 prompts across all possible combinations of capital letters as action labels (325 total combinations). For each combination, we calculate the KL divergence between the target and extracted output distributions, averaged across the 5 prompts, and select the combination with the lowest divergence. The resulting action labels are shown in Table~\ref{tab:action_labels}.

The extracted initial probabilities for each pair of action labels in Table~\ref{tab:action_labels} are shown in Table~\ref{tab:opp_init_all}, with separate results for the IPD (\subref{tab:opp_initIPD}), IMP (\subref{tab:opp_initIMP}), and ICG (\subref{tab:opp_initC}).

\begin{table}[!htbp]
\centering
\caption{Naive learner's initial action probabilities across three games (IPD, IMP, ICG) under varying action labels for the 5 distinct prompts encountered during training.}
\label{tab:opp_init_all}

\begin{subtable}{\textwidth}
\centering
\caption{Initial cooperation probability in the IPD.}
\label{tab:opp_initIPD}
\begin{tabular}{lccccc}
\toprule
\multicolumn{1}{c}{\bf Action Labels} &
\multicolumn{1}{c}{\bf $\mathbf{p^0_\text{NL}(\text{C})}$} &
\multicolumn{1}{c}{\bf $\mathbf{p^0_\text{NL}(\text{C} \mid \text{CC})}$} &
\multicolumn{1}{c}{\bf $\mathbf{p^0_\text{NL}(\text{C} \mid \text{CD})}$} &
\multicolumn{1}{c}{\bf $\mathbf{p^0_\text{NL}(\text{C} \mid \text{DC})}$} &
\multicolumn{1}{c}{\bf $\mathbf{p^0_\text{NL}(\text{C} \mid \text{DD})}$} \\
\midrule
$w_{a1}=\text{H}, w_{a2}=\text{K}$ & 0.60 & 0.89 & 0.89 & 0.70 & 0.68 \\
$w_{a1}=\text{N}, w_{a2}=\text{Y}$ & 0.68 & 0.38 & 0.87 & 0.58 & 0.77 \\
$w_{a1}=\text{I}, w_{a_2}=\text{X}$ & 0.12 & 0.21 & 0.75 & 0.34 & 0.24 \\
\bottomrule
\end{tabular}
\end{subtable}

\vspace{1em}

\begin{subtable}{\textwidth}
\centering
\caption{Initial probability of playing heads in the IMP.}
\label{tab:opp_initIMP}
\begin{tabular}{lccccc}
\toprule
\multicolumn{1}{c}{\bf Action Labels} &
\multicolumn{1}{c}{\bf $\mathbf{p^0_\text{NL}(\text{H})}$} &
\multicolumn{1}{c}{\bf $\mathbf{p^0_\text{NL}(\text{H} \mid \text{HH})}$} &
\multicolumn{1}{c}{\bf $\mathbf{p^0_\text{NL}(\text{H} \mid \text{HT})}$} &
\multicolumn{1}{c}{\bf $\mathbf{p^0_\text{NL}(\text{H} \mid \text{TH})}$} &
\multicolumn{1}{c}{\bf $\mathbf{p^0_\text{NL}(\text{H} \mid \text{TT})}$} \\
\midrule
$w_{a1}=\text{S}, w_{a2}=\text{Y}$ & 0.71 & 0.78 & 0.87 & 0.57 & 0.70 \\
$w_{a1}=\text{N}, w_{a2}=\text{Y}$ & 0.48 & 0.32 & 0.60 & 0.56 & 0.66 \\
$w_{a1}=\text{N}, w_{a_2}=\text{M}$ & 0.30 & 0.20 & 0.36 & 0.13 & 0.18 \\
\bottomrule
\end{tabular}
\end{subtable}

\vspace{1em}

\begin{subtable}{\textwidth}
\centering
\caption{Initial swerving probability in the ICG.}
\label{tab:opp_initC}
\begin{tabular}{lccccc}
\toprule
\multicolumn{1}{c}{\bf Action Labels} &
\multicolumn{1}{c}{\bf $\mathbf{p^0_\text{NL}(\text{S})}$} &
\multicolumn{1}{c}{\bf $\mathbf{p^0_\text{NL}(\text{S} \mid \text{SS})}$} &
\multicolumn{1}{c}{\bf $\mathbf{p^0_\text{NL}(\text{S} \mid \text{SG})}$} &
\multicolumn{1}{c}{\bf $\mathbf{p^0_\text{NL}(\text{S} \mid \text{GS})}$} &
\multicolumn{1}{c}{\bf $\mathbf{p^0_\text{NL}(\text{S} \mid \text{GG})}$} \\
\midrule
$w_{a1}=\text{T}, w_{a2}=\text{K}$ & 0.61 & 0.90 & 0.87 & 0.82 & 0.82 \\
$w_{a1}=\text{N}, w_{a2}=\text{M}$ & 0.41 & 0.40 & 0.71 & 0.37 & 0.61 \\
$w_{a1}=\text{T}, w_{a_2}=\text{F}$ & 0.24 & 0.47 & 0.47 & 0.02 & 0.11 \\
\bottomrule
\end{tabular}
\end{subtable}

\end{table}


\subsection{Shaper evaluation for varying game lengths for the IPD, IMP, and ICG}
\label{app:different_T}

Table~\ref{tab:differnt_T} presents the evaluation results for the shapers trained in Section~\ref{sec:main_results} for varying game lengths ($T=50$, $T=100$). The results demonstrate that shaper performance remains consistent across different game lengths, with no significant degradation in exploitation capability as episodes become longer.

\begin{table}[h]
\caption{Post-training evaluation results for the IPD, IMP, and ICG comparing baseline (two naive learners) versus shaper-naive learner pairs for varying game lengths ($T= 20, 50, 100)$. Average rewards per step are reported with 95\% confidence intervals across 5 random seeds. Illegal actions are excluded from the analysis (comprising 2\% of actions in IPD, 0.1\% in IMP, and 1\% in ICG).}
\label{tab:differnt_T}
\begin{center}
\begin{tabular}{lcccccc}
\toprule
& \multicolumn{2}{c}{\bf IPD} & \multicolumn{2}{c}{\bf IMP} & \multicolumn{2}{c}{\bf ICG}\\
\cmidrule(lr){2-3} \cmidrule(lr){4-5} \cmidrule(lr){6-7}
&Shaper & Opponent & Shaper & Opponent & Shaper & Opponent\\
\midrule
$T=20$ & $3.96 \pm 0.01$ & $0.10 \pm 0.04$ & $0.99 \pm 0.01$ & $-0.99 \pm 0.01$ & $2.98 \pm 0.01$ & $1.01 \pm 0.01$\\
$T=50$ & $3.97 \pm 0.01$ & $0.09 \pm 0.04$ & $0.99 \pm 0.01$ & $-0.99 \pm 0.01$ & $2.99 \pm 0.01$ & $1.01 \pm 0.01$\\
$T=100$ & $3.97 \pm 0.01$ & $0.09 \pm 0.04$ & $0.99 \pm 0.01$ & $-0.99 \pm 0.01$ & $2.99 \pm 0.00$ & $1.01 \pm 0.01$\\
\bottomrule
\end{tabular}
\end{center}
\end{table}

\subsection{Training Dynamics}
\label{app:training_dynamics}

\begin{figure}[h]
\begin{center}
\includegraphics{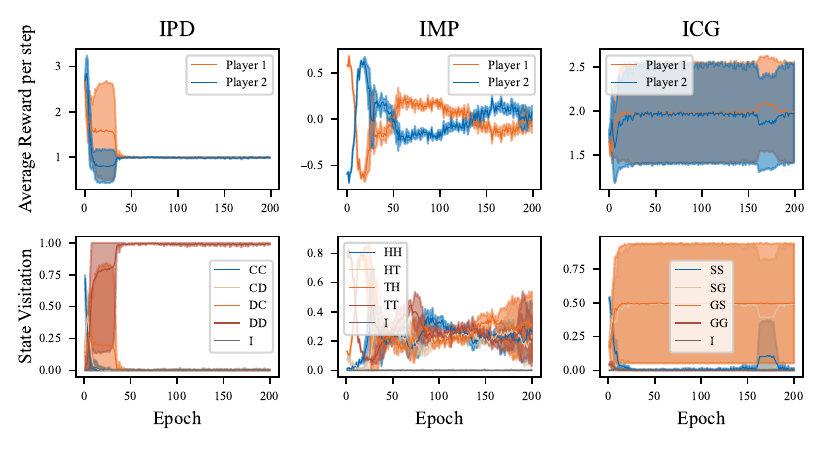}
\end{center}
\caption{Average reward per step (top row) and state visitation (bottom row) during training for the baseline experiments across the IPD, IMP and ICG reported in Section~\ref{sec:main_results}. In the state visitation figures, the outcome ``I" encompasses all transitions where either player
chose $a_\text{null}$. The results are reported along with a 95\% confidence interval over 5 random seeds, except for the ICG experiment, for which we use 10 seeds.}
\label{fig:baseline_plots}
\end{figure}

\begin{figure}[h]
\begin{center}
\includegraphics{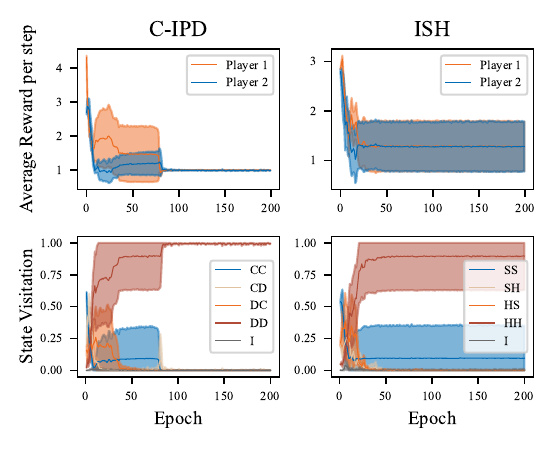}
\end{center}
\caption{Average reward per step (top row) and state visitation (bottom row) during training for the baseline experiments across the C-IPD and ISH reported in Section~\ref{sec:shaping_cooperative}. In the state visitation figures, the outcome ``I" encompasses all transitions where either player
chose $a_\text{null}$. The results are reported along with a 95\% confidence interval over 10 random seeds.}
\label{fig:cooperative_shaper}
\end{figure}

\begin{figure}[h]
\begin{center}
\includegraphics{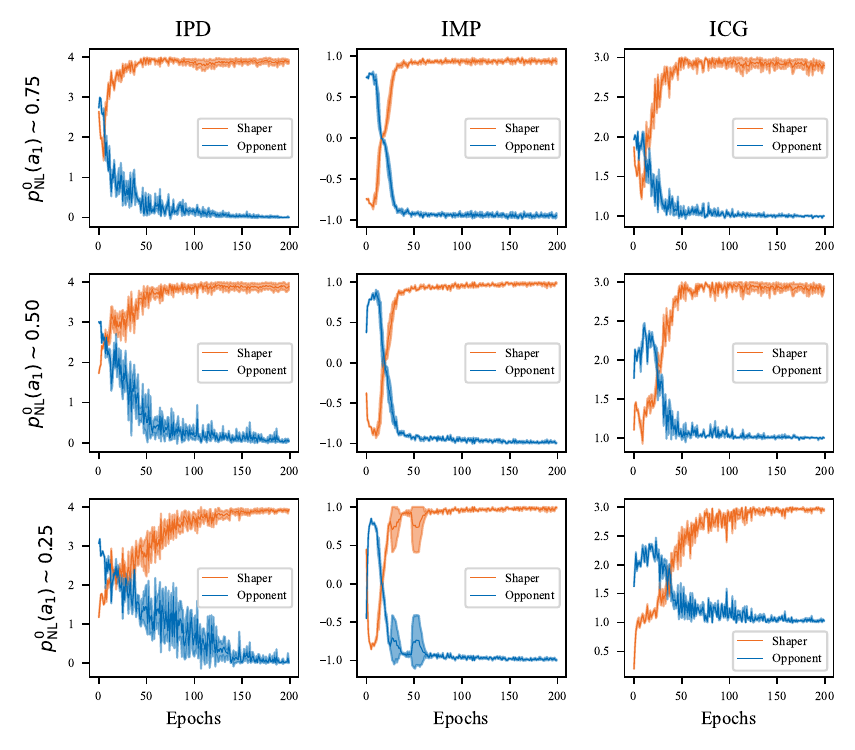}
\end{center}
\caption{Average reward per step during training for the shaping experiments across the IPD, IMP, and ICG with different opponent initializations (Section~\ref{sec:main_results_robust}). Each row corresponds to an opponent with a different initial probability of generating action $a_1$ (``Cooperate" in the IPD, ``Heads" in the IMP, and ``Swerve" in the ICG). The results are reported along with a 95\% confidence interval over 5 random seeds.}
\label{fig:shaper_diff_opp_r}
\end{figure}

\begin{figure}[h]
\begin{center}
\includegraphics{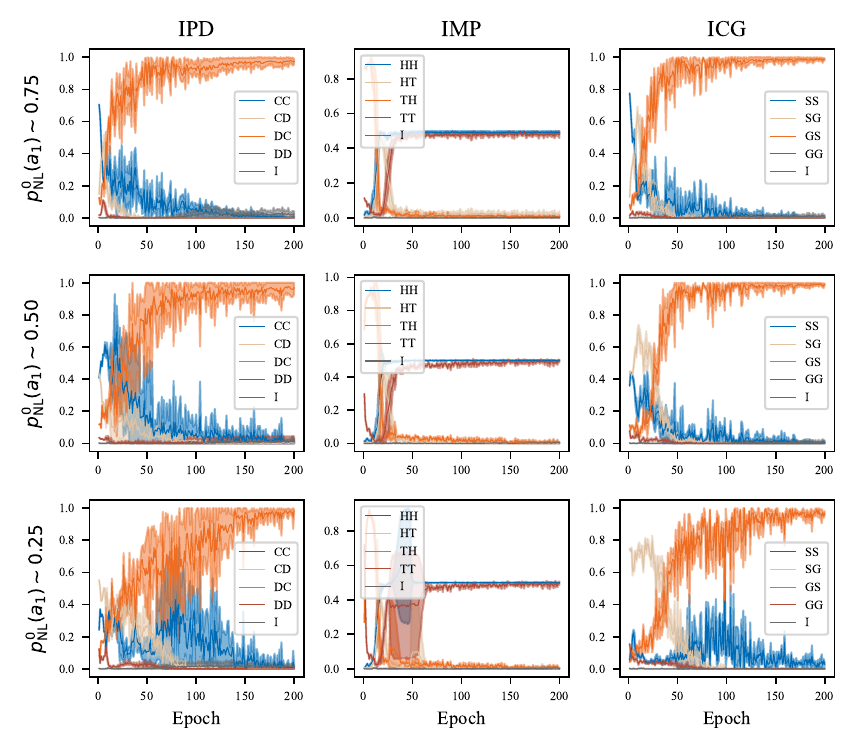}
\end{center}
\caption{State visitation during training for the shaping experiments across the IPD, IMP, and ICG with different opponent initializations (Section~\ref{sec:main_results_robust}). Each row corresponds to an opponent with a different initial probability of generating action $a_1$ (``Cooperate" in the IPD, ``Heads" in the IMP, and ``Swerve" in the ICG). For all games, the outcome ``I" encompasses all transitions where either player
chose $a_\text{null}$. The results are reported along with a 95\% confidence interval over 5 random seeds.}
\label{fig:shaper_diff_opp_state_visitations}
\end{figure}

In this section we present the average reward per step and state visitation throughout training for various experiments discussed in the main text. These figures complement the evaluation results by showing how agent behaviors evolved during the learning process. Figure~\ref{fig:baseline_plots} presents the training dynamics for the baseline experiments across the IPD, IMP and ICG (Section~\ref{sec:main_results}). Figure~\ref{fig:cooperative_shaper} shows the baseline training dynamics for the cooperative shaping experiments (Section~\ref{sec:shaping_cooperative}). Finally, Figures~\ref{fig:shaper_diff_opp_r} and \ref{fig:shaper_diff_opp_state_visitations} present the training dynamics for shaper experiments against three different opponent types across the IPD, IMP, and ICG, with evaluation results reported in Section~\ref{sec:main_results_robust}.

\clearpage 
\subsection{Training Prompts}
\label{app:training_prompts}

We use three types of prompts throughout training: the base prompt, the state-only prompt, and the state occurrence prompt. 

\begin{figure}[h]
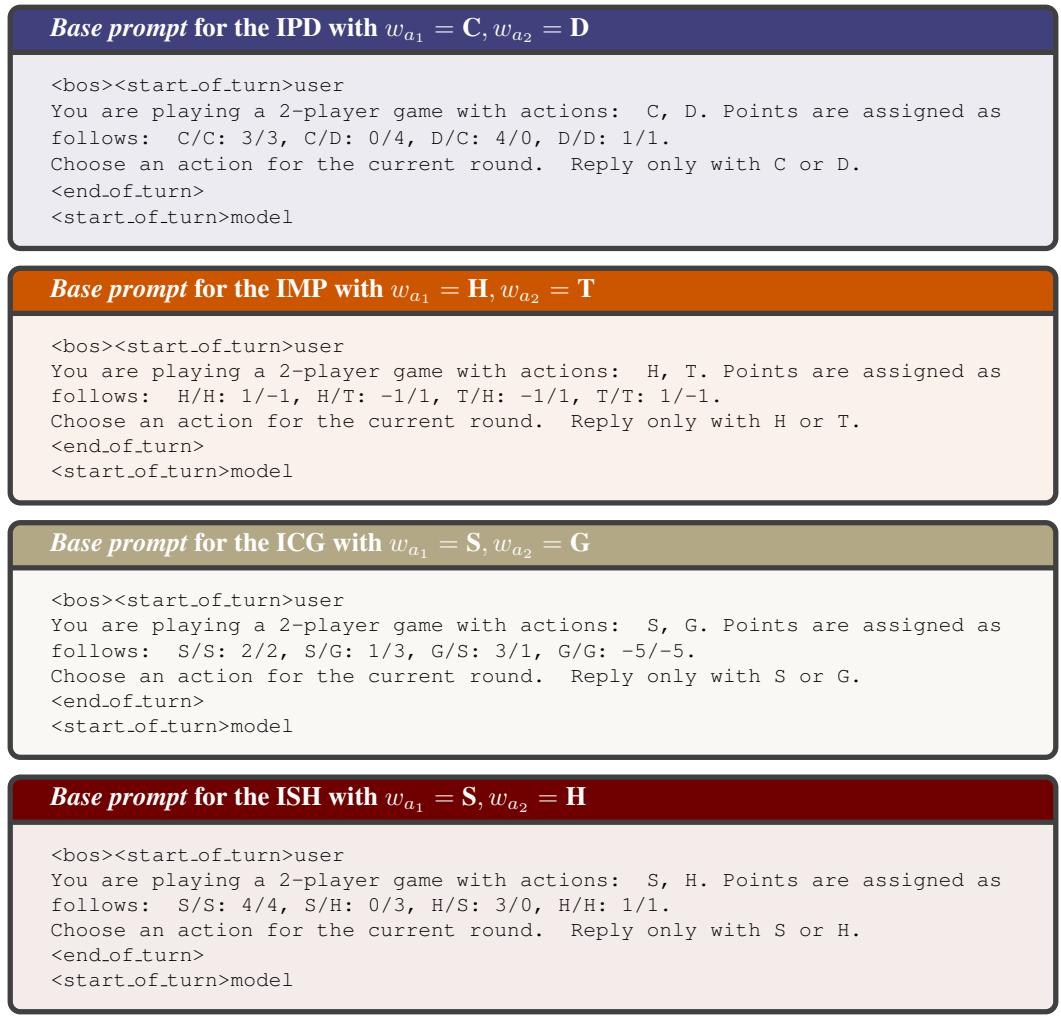

\begin{center}
\begin{tcolorbox}[
    colback=lightblue,
    colframe=darkgray,
    fonttitle=\bfseries\color{white},
    fontupper=\ttfamily\fontsize{8pt}{10pt}\selectfont, 
    title={\textit{Base prompt} for the IPD with $w_{a_1} = \text{C}, w_{a_2} = \text{D}$},
    coltitle=white,
    colbacktitle=darkblue,
    rounded corners,
    boxrule=2pt,
    width=\textwidth 
]
\texttt{<bos><start\_of\_turn>user}

\texttt{You are playing a 2-player game with actions:  C, D. Points are assigned as follows: C/C: 3/3,  C/D: 0/4, D/C: 4/0, D/D: 1/1.}

\texttt{Choose an action for the current round. Reply only with C or D.}

\texttt{<end\_of\_turn>}

\texttt{<start\_of\_turn>model}
\end{tcolorbox}
\begin{tcolorbox}[
    colback=lightburntorange,
    colframe=darkgray,
    fonttitle=\bfseries\color{white},
    fontupper=\ttfamily\fontsize{8pt}{9.5pt}\selectfont, 
    title={\textit{Base prompt }for the IMP with $w_{a_1} = \text{H}, w_{a_2} = \text{T}$},
    coltitle=white,
    colbacktitle=burntorange,
    rounded corners,
    boxrule=2pt,
    width=\textwidth 
]
\texttt{<bos><start\_of\_turn>user}

\texttt{You are playing a 2-player game with actions:  H, T. Points are assigned as follows: H/H: 1/-1,  H/T: -1/1, T/H: -1/1, T/T: 1/-1.}

\texttt{Choose an action for the current round. Reply only with H or T.}

\texttt{<end\_of\_turn>}

\texttt{<start\_of\_turn>model}
\end{tcolorbox}
\begin{tcolorbox}[
    colback=lightcream,
    colframe=darkgray,
    fonttitle=\bfseries\color{white},
    fontupper=\ttfamily\fontsize{8pt}{9.5pt}\selectfont, 
    title={\textit{Base prompt }for the ICG with $w_{a_1} = \text{S}, w_{a_2} = \text{G}$},
    coltitle=white,
    colbacktitle=cream,
    rounded corners,
    boxrule=2pt,
    width=\textwidth 
]
\texttt{<bos><start\_of\_turn>user}

\texttt{You are playing a 2-player game with actions:  S, G. Points are assigned as follows: S/S: 2/2,  S/G: 1/3, G/S: 3/1, G/G: -5/-5.}

\texttt{Choose an action for the current round. Reply only with S or G.}

\texttt{<end\_of\_turn>}

\texttt{<start\_of\_turn>model}
\end{tcolorbox}

\begin{tcolorbox}[
    colback=lightmaroon,
    colframe=darkgray,
    fonttitle=\bfseries\color{white},
    fontupper=\ttfamily\fontsize{8pt}{9.5pt}\selectfont, 
    title={\textit{Base prompt }for the ISH with $w_{a_1} = \text{S}, w_{a_2} = \text{H}$},
    coltitle=white,
    colbacktitle=maroon,
    rounded corners,
    boxrule=2pt,
    width=\textwidth 
]
\texttt{<bos><start\_of\_turn>user}

\texttt{You are playing a 2-player game with actions:  S, H. Points are assigned as follows: S/S: 4/4,  S/H: 0/3, H/S: 3/0, H/H: 1/1.}

\texttt{Choose an action for the current round. Reply only with S or H.}

\texttt{<end\_of\_turn>}

\texttt{<start\_of\_turn>model}
\end{tcolorbox}
\end{center}
\caption{Base prompts for the IPD, IMP, ICG and ISH. The structure remains the same across games, with the only differences being the action labels and reward matrices.}
\label{fig:base_prompts}
\end{figure}

The \textit{base prompt} (Figure~\ref{fig:base_prompts}) includes a game description specifying the number of players, available actions, and payoff structure. It also contains an instruction to select an action using the designated action labels $w_{a_1}, w_{a_2}$.

The \textit{state-only prompt} (Figure~\ref{fig:state_only_prompts}) includes all information from the base prompt and additionally incorporates the last joint action of both players. Naive learners receive this prompt for all game rounds except the first, where they receive the base prompt. Shapers receive this prompt only at the second round of each trial, when one previous action has been played.

Lastly, the \textit{state occurrence prompt} (Figure~\ref{fig:state_occ_prompts}) includes all information from the state-only prompt and additionally incorporates the state counts for all previous rounds within the current trial. Shaping agents receive this prompt for all interactions beyond the first two rounds. The prompt is dynamically updated as follows: when new actions are played, the state count is updated to include the most recent state, and the new joint action is displayed in the current state field.

\begin{figure}[h]
\begin{center}
\begin{tcolorbox}[
    colback=lightblue,
    colframe=darkgray,
    fonttitle=\bfseries\color{white},
    fontupper=\ttfamily\fontsize{8pt}{10pt}\selectfont, 
    title={Example of \textit{state-only prompt} for the IPD with $w_{a_1} = \text{C}, w_{a_2} = \text{D}$},
    coltitle=white,
    colbacktitle=darkblue,
    rounded corners,
    boxrule=2pt,
    width=\textwidth 
]
\texttt{<bos><start\_of\_turn>user}

\texttt{You are playing a 2-player game with actions:  C, D. Points are assigned as follows: C/C: 3/3,  C/D: 0/4, D/C: 4/0, D/D: 1/1.}

\texttt{<STATE>In the previous round, you played C and your opponent played C.}

\texttt{Choose an action for the current round. Reply only with C or D.}

\texttt{<end\_of\_turn>}

\texttt{<start\_of\_turn>model}
\end{tcolorbox}
\begin{tcolorbox}[
    colback=lightburntorange,
    colframe=darkgray,
    fonttitle=\bfseries\color{white},
    fontupper=\ttfamily\fontsize{8pt}{10pt}\selectfont, 
    title={Example of \textit{state-only prompt} for the IMP with $w_{a_1} = \text{H}, w_{a_2} = \text{T}$},
    coltitle=white,
    colbacktitle=burntorange,
    rounded corners,
    boxrule=2pt,
    width=\textwidth 
]
\texttt{<bos><start\_of\_turn>user}

\texttt{You are playing a 2-player game with actions:  H, T. Points are assigned as follows: H/H: 1/-1,  H/T: -1/1, T/H: -1/1, T/T: 1/-1.}

\texttt{<STATE>In the previous round, you played H and your opponent played H.}

\texttt{Choose an action for the current round. Reply only with H or T.}

\texttt{<end\_of\_turn>}

\texttt{<start\_of\_turn>model}
\end{tcolorbox}
\begin{tcolorbox}[
    colback=lightcream,
    colframe=darkgray,
    fonttitle=\bfseries\color{white},
    fontupper=\ttfamily\fontsize{8pt}{9.5pt}\selectfont, 
    title={Example of \textit{state-only prompt} for the ICG with $w_{a_1} = \text{S}, w_{a_2} = \text{G}$},
    coltitle=white,
    colbacktitle=cream,
    rounded corners,
    boxrule=2pt,
    width=\textwidth 
]
\texttt{<bos><start\_of\_turn>user}

\texttt{You are playing a 2-player game with actions:  S, G. Points are assigned as follows: S/S: 2/2,  S/G: 1/3, G/S: 3/1, G/G: -5/-5.}

\texttt{<STATE>In the previous round, you played S and your opponent played S.}

\texttt{Choose an action for the current round. Reply only with S or G.}

\texttt{<end\_of\_turn>}

\texttt{<start\_of\_turn>model}
\end{tcolorbox}

\begin{tcolorbox}[
    colback=lightmaroon,
    colframe=darkgray,
    fonttitle=\bfseries\color{white},
    fontupper=\ttfamily\fontsize{8pt}{9.5pt}\selectfont, 
    title={Example of \textit{state-only prompt} for the ISH with $w_{a_1} = \text{S}, w_{a_2} = \text{G}$},
    coltitle=white,
    colbacktitle=maroon,
    rounded corners,
    boxrule=2pt,
    width=\textwidth 
]
\texttt{<bos><start\_of\_turn>user}

\texttt{You are playing a 2-player game with actions:  S, H. Points are assigned as follows: S/S: 4/4,  S/H: 0/3, H/S: 3/0, H/H: 1/1.}

\texttt{<STATE>In the previous round, you played S and your opponent played S.}

\texttt{Choose an action for the current round. Reply only with S or H.}

\texttt{<end\_of\_turn>}

\texttt{<start\_of\_turn>model}
\end{tcolorbox}
\end{center}
\caption{Example state-only prompts for the IPD, IMP, ICG and ISH. The structure remains the same across games, with the only differences being the action labels and reward matrices. The specific examples shown are for rounds in which the previous joint action is $(a_1, a_1)$.}
\label{fig:state_only_prompts}
\end{figure}

\begin{figure}[h]
\begin{center}
\begin{tcolorbox}[
    colback=lightblue,
    colframe=darkgray,
    fonttitle=\bfseries\color{white},
    fontupper=\ttfamily\fontsize{8pt}{10pt}\selectfont, 
    title={Example of \textit{state occurrence prompt} for the IPD with $w_{a_1} = \text{C}, w_{a_2} = \text{D}$},
    coltitle=white,
    colbacktitle=darkblue,
    rounded corners,
    boxrule=2pt,
    width=\textwidth 
]
\texttt{<bos><start\_of\_turn>user}

\texttt{You are playing a 2-player game with actions:  C, D. Points are assigned as follows: C/C: 3/3,  C/D: 0/4, D/C: 4/0, D/D: 1/1.}

\texttt{<ADDITIONAL INFORMATION>The occurrence of each state in the current game has been CC:0, CD:0, DC:0, DD:5.}

\texttt{<STATE>In the previous round, you played C and your opponent played C.}

\texttt{Choose an action for the current round. Reply only with C or D.}

\texttt{<end\_of\_turn>}

\texttt{<start\_of\_turn>model}
\end{tcolorbox}

\begin{tcolorbox}[
    colback=lightburntorange,
    colframe=darkgray,
    fonttitle=\bfseries\color{white},
    fontupper=\ttfamily\fontsize{8pt}{10pt}\selectfont, 
    title={Example of \textit{state occurrence prompt} for the IMP with $w_{a_1} = \text{H}, w_{a_2} = \text{T}$},
    coltitle=white,
    colbacktitle=burntorange,
    rounded corners,
    boxrule=2pt,
    width=\textwidth 
]
\texttt{<bos><start\_of\_turn>user}

\texttt{You are playing a 2-player game with actions:  H, T. Points are assigned as follows: H/H: 1/-1,  H/T: -1/1, T/H: -1/1, T/T: 1/-1.}

\texttt{<ADDITIONAL INFORMATION>The occurrence of each state in the current game has been HH:0, HT:0, TH:0, TT:5.}

\texttt{<STATE>In the previous round, you played H and your opponent played H.}

\texttt{Choose an action for the current round. Reply only with H or T.}

\texttt{<end\_of\_turn>}

\texttt{<start\_of\_turn>model}
\end{tcolorbox}

\begin{tcolorbox}[
    colback=lightcream,
    colframe=darkgray,
    fonttitle=\bfseries\color{white},
    fontupper=\ttfamily\fontsize{8pt}{9.5pt}\selectfont, 
    title={Example of \textit{state occurrence prompt} for the ICG with $w_{a_1} = \text{S}, w_{a_2} = \text{G}$},
    coltitle=white,
    colbacktitle=cream,
    rounded corners,
    boxrule=2pt,
    width=\textwidth 
]
\texttt{<bos><start\_of\_turn>user}

\texttt{You are playing a 2-player game with actions:  S, G. Points are assigned as follows: S/S: 2/2,  S/G: 1/3, G/S: 3/1, G/G: -5/-5.}

\texttt{<ADDITIONAL INFORMATION>The occurrence of each state in the current game has been SS:0, SG:0, GS:0, GG:5.}

\texttt{<STATE>In the previous round, you played S and your opponent played S.}

\texttt{Choose an action for the current round. Reply only with S or G.}

\texttt{<end\_of\_turn>}

\texttt{<start\_of\_turn>model}
\end{tcolorbox}

\begin{tcolorbox}[
    colback=lightmaroon,
    colframe=darkgray,
    fonttitle=\bfseries\color{white},
    fontupper=\ttfamily\fontsize{8pt}{9.5pt}\selectfont, 
    title={Example of \textit{state occurrence prompt} for the ISH with $w_{a_1} = \text{S}, w_{a_2} = \text{H}$},
    coltitle=white,
    colbacktitle=maroon,
    rounded corners,
    boxrule=2pt,
    width=\textwidth 
]
\texttt{<bos><start\_of\_turn>user}

\texttt{You are playing a 2-player game with actions:  S, H. Points are assigned as follows: S/S: 4/4,  S/H: 0/3, H/S: 3/0, H/H: 1/1.}

\texttt{<ADDITIONAL INFORMATION>The occurrence of each state in the current game has been SS:0, SH:0, HS:0, HH:5.}

\texttt{<STATE>In the previous round, you played S and your opponent played S.}

\texttt{Choose an action for the current round. Reply only with S or H.}

\texttt{<end\_of\_turn>}

\texttt{<start\_of\_turn>model}
\end{tcolorbox}

\end{center}
\caption{Example state-occurrence prompts for the IPD, IMP, ICG and ISH. The structure remains the same across games, with the only differences being the action labels and reward matrices.  The specific examples shown are for rounds in which all previous joint actions within the game are $(a_2, a_2)$, except the last one, which is $(a_1, a_1)$.}
\label{fig:state_occ_prompts}
\end{figure}

\end{document}